\documentclass[11pt]{article}

\usepackage[final]{acl}

\usepackage{times}
\usepackage{latexsym}

\usepackage[T1]{fontenc}

\usepackage[utf8]{inputenc}

\usepackage{microtype}

\usepackage{inconsolata}

\usepackage{graphicx}

\usepackage{amsmath}
\usepackage{algorithm}
\usepackage{algpseudocode}
\usepackage{xcolor}
\usepackage{multirow}
\usepackage{xurl}

\title{Expert Preference-based Evaluation of Automated Related Work Generation}

\author{Furkan \c{S}ahinu\c{c}$^{1,2}$, Subhabrata Dutta$^{1}$, Iryna Gurevych$^{1,2}$ \\
$^{1}$Ubiquitous Knowledge Processing Lab (UKP Lab) \\
Department of Computer Science and Hessian Center for AI (hessian.AI) \\
Technical University of Darmstadt \\
$^{2}$Konrad Zuse School of Excellence in Learning and Intelligent Systems (ELIZA) \\
\url{www.ukp.tu-darmstadt.de}}

\newcommand{\framework}{{\tt GREP}}

\begin{document}
\maketitle
\begin{abstract}
Expert domain writing, such as scientific writing, typically demands extensive domain knowledge. Although large language models (LLMs) show promising potential in this task, evaluating the quality of automatically generated scientific writing is a crucial open issue, as it requires knowledge of domain-specific criteria and the ability to discern expert preferences. Conventional automatic evaluation metrics and LLM-as-a-judge systems, primarily designed for mainstream NLP tasks, are insufficient to grasp expert preferences and domain-specific quality standards. To address this gap and support realistic human-AI collaborative writing, we focus on related work generation, one of the most challenging scientific tasks, as an exemplar. We propose \framework, a multi-turn evaluation framework that integrates classical related work evaluation criteria with expert-specific preferences. \framework\ decomposes the evaluation into smaller fine-grained dimensions. This localized evaluation is further augmented with contrastive examples to provide detailed contextual guidance for the evaluation dimensions. Empirical investigation reveals that \framework\ is able to assess the quality of related work sections in a much more robust manner compared to standard LLM judges, reflects natural scenarios of scientific writing, and bears a strong correlation with the assessment of human experts. We also observe that generations from widely used LLMs struggle to satisfy validation constraints of a suitable related work section. We make our code\footnote{Code: \url{https://github.com/UKPLab/eacl2027-expert-eval-rw}} and data\footnote{Data: \url{https://tudatalib.ulb.tu-darmstadt.de/handle/tudatalib/4700}} publicly available.

\end{abstract}

\section{Introduction}

With the advent of Large Language Models (LLMs) and Large Reasoning Models (LRMs), there has been an increasing attempt to incorporate AI assistance in expert domain problems, such as scientific writing \cite{Salvagno:2023,Wang:2024AutoSurvey,Lin:2025}. As opposed to commonplace text generation tasks \cite{Dong:2022}, such tasks require vast domain knowledge \cite{Evans:1995}. The AI agent should reason over novel information in relation to the domain knowledge \cite{Wen:2024} and should cater to the preferences of a human expert in a meaningful way \cite{Dutta:2025,Gao:2024,Aroca-ouellette:2025}. This phenomenon is also valid while evaluating generated artifacts, as assessing generated text has long been challenging \cite{Gehrmann:2023} due to the possibility of numerous valid generations differing in surface-level lexicons (similarly, incorrect generations sharing similar lexical traits with a correct one). Unlike tasks with formally verifiable answers, such as mathematical reasoning \cite{Hendrycks:2021} and code \cite{Chen:2021Eval}, this difficulty increases in scientific writing, which often requires expert judgment rather than simple or automatic verification.

\begin{figure*}[ht]
\centering
\includegraphics[width=1\linewidth]{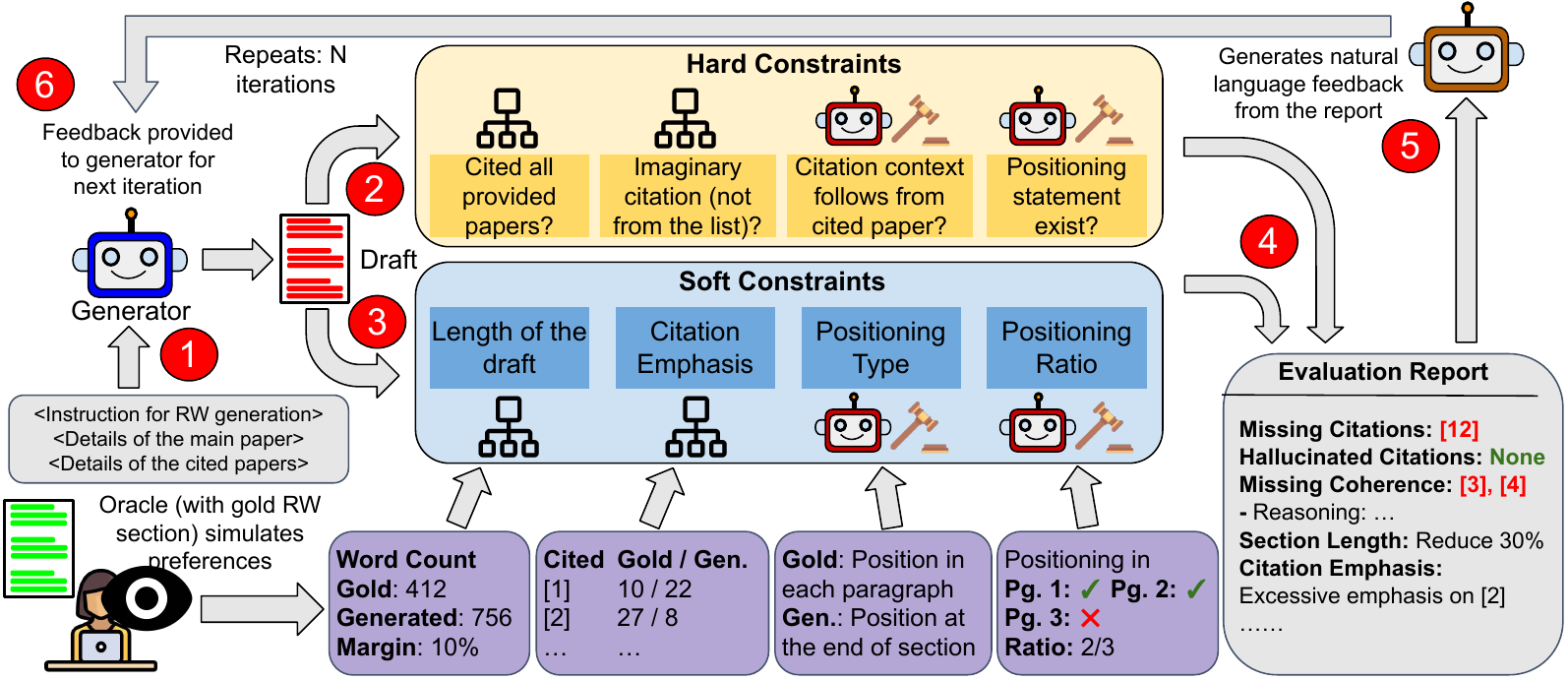}
\caption{Illustrative description of \framework. Generated related work drafts are evaluated by dedicated modules that consider hard and soft constraints. Oracle with access to the gold RW section defines the preferences over soft constraints, which are compared against the generated draft. Natural language feedback is generated based on the evaluation report to guide the generator LLM in producing the revised draft in the next iteration. Step numbers indicate the direction of the loop.}
\label{fig:framework}
\end{figure*}

Leveraging LLMs' natural language understanding, the LLM-as-a-judge paradigm \cite{Liu:2023,Zheng:2023} has emerged as a partial solution by scoring or comparing candidate generations. However, our own experiments, along with several recent investigations \cite{Gao:2025,Li:2024NLG,Szymanski:2025}, highlight the key limitations of these judge models: pretraining biases, inability to perform domain-grounded reasoning, misalignment with expert preferences, and lack of transparency in judgment. Such frameworks lack the knowledge of \textit{what to judge} and \textit{how to judge} especially for scientific writing tasks.

In this work, we focus on a critical component of the scientific writing pipeline: generating the Related Work (RW) section of a paper given a list of relevant papers to be cited. Following \cite{Dutta:2025}, we adopt the view that RW generation requires collaboration between the AI agent and the human expert, and subsequently, the utility of the solution should reflect the expert's preferences. This leads to our main research question: How can we evaluate an LLM’s ability to generate and refine a related work section? We embed this evaluation in a multi-turn generation setup, where the generator (i.e., system under evaluation) iteratively refines the generated draft upon feedback from the evaluation of the prior iteration.

\paragraph{Contributions and findings.} Our contribution lies in evaluation methodology and benchmarking rather than in a new generation model. To this end, we initially construct a new RW generation dataset with rich contextual information by addressing the limitations\footnote{Typically providing only Titles and Abstracts; we include Introduction sections collected from heterogeneous sources.} of the previous datasets ({\bf C1}). 
We introduce a fine-grained RW evaluation rubric ({\bf C2}) where hard constraints (i.e., requirements for being a valid RW section) complement soft constraints (i.e., reflecting human preferences over multiple valid RW sections, e.g., emphasis on certain cited papers). 
We design \framework\ (\textbf{G}ranular \textbf{R}elated-work \textbf{E}valuation based on \textbf{P}references), a multi-turn evaluation system (see Figure~\ref{fig:framework} for outlined operation) to assess both the quality of generated RW sections and the generator's ability to incorporate evaluation feedback ({\bf C3}). We provide two variants of \framework\ to enhance accessibility for the community: Precise\framework\ which uses proprietary LLM judges (higher cost, higher accuracy) and Open\framework, a cost-efficient version, which relies on open weight models. \framework\ unifies both deterministically verifiable criteria and criteria requiring deeper natural language understanding. For the latter, motivated by the limitations of existing LLM-as-a-judge systems, we redesign the evaluation based on two principles: 
i) {\em Localized judgment}, where we specify the precise evaluation context (e.g., whether a citation context aligns with the cited paper) rather than holistic evaluation, addressing the \textit{what to judge} problem. Decomposition of the complex evaluation task into multiple simpler, semi-objective tests improves transparency of \framework.
ii) {\em Manipulated contrastive examples}, supplied in context to inform the model of the judgment distribution, addressing the \textit{how to judge problem}.
We validate \framework\ via an expert study ({\bf C4}): 16 domain experts are asked to independently evaluate LLM-generated RW sections in a pairwise manner with multi-turn interactions and select the winning RW generators. While specialized LLM judges deliver subpar matching with expert judgments (e.g., 53\% match in citation faithfulness), assessments from Precise\framework\ 
and Open\framework\ provide judgments that are closely similar to experts (e.g., \textbf{78\%} and \textbf{66\%} matching in citation faithfulness). 
Finally, we use \framework\ to shed light on current LLMs' capability to generate RW sections. They struggle to faithfully cite prior work ({\bf F1}) -- the best performing model, o3-mini, could only do it 20\% of the time. Improvement upon explicit feedback is rare; failure modes can be associated with i) struggling to keep track of multiple improvement aspects presented in the feedback, ii) introducing inconsistent edits that worsen pre-feedback quality, and iii) inability to incorporate even simple preference-based instructions like adjusting the length of the generated RW section ({\bf F2}). A severe decline in quality is observed when user preferences change midway through the interaction, e.g., introducing new papers or changing section organization midway through the interaction ({\bf F3}).

\section{Related Work}

\paragraph{Automated Related Work generation.} Before the LLM era, citation text or RW generation tasks were mainly framed as summarization task, addressed by different model architectures designed around specific input-output configurations \cite{Yasunaga:2019,Xing:2020,Lu:2020,Luu:2021,Ge:2021,Li:2022,Liu:2023Causal,Chen:2021,Chen:2022}. The flexibility of LLMs in performing complex tasks has enabled the use of diverse inputs, such as citation intent or citation spans \cite{Arita:2022,Jung:2022,Martinboyle:2024,Sahinuc:2024,Li:2025Explaining}. This capability is not limited to the use of different input configurations, but has also led to the development of agentic or tool-augmented pipelines to implement different steps in the literature review writing process such as paper retrieval and outline of ideas \cite{Shi:2023,Wang:2024AutoSurvey,Agarwal:2025,Liang:2025,Wang:2025,Liu:2025Select}. Furthermore, recent frameworks for human-AI collaboration leveraging natural language interactions have also been proposed for related work generation \cite{Shao:2025}. However, most of these works (1) rely mainly on titles and abstracts of cited papers, which limits the context available for generating RW sections, and (2) use evaluation schemes that do not consider expert preferences required to distinguish high-quality RWs containing domain-specific nuances, such as the position of the paper among the previous literature or the emphasis of each cited paper. In contrast, we extend the context with introduction sections of cited papers and consider expert preferences in both the generation and evaluation phases.

\paragraph{Evaluation of AI-generated content.}
Evaluating natural language generation is inherently challenging \cite{Gehrmann:2023}, particularly for tasks with multiple valid outputs requiring expert domain knowledge, such as RW generation \cite{Li:2024,Sahinuc:2024}. Traditional metrics like ROUGE \cite{Lin:2004} and BERTScore \cite{Zhang:2020} are unable to consider expert domain requirements \cite{Nimah:2023}. LLM-as-a-judge methods have been proposed as a remedy due to their potential to serve as a flexible and versatile evaluation system \cite{Liu:2023,Zheng:2023}. However, LLM evaluators have been shown to lack robust performance \cite{Gao:2025,Li:2024NLG,Szymanski:2025} as they can show bias towards specific positions in comparisons \cite{Wang:2024Fair} or prefer longer responses \cite{Zheng:2023}. To better align with human judgments, checklist-based evaluation systems have been proposed to assess whether models meet task-specific criteria \cite{Pereira:2024,Lee:2025,Que:2024,Li:2025}. However, such checklists (machine-generated or human-curated) are designed for task-level applicability, whereas expert tasks demand unique, instance-specific criteria reflecting individual expert preferences. Moreover, framing checklist evaluation as a binary QA task \cite{Qin:2024} is insufficient, since it lacks the required context for iterative co-construction. \citet{Jourdan:2025} also suggest that LLM-as-a-judge should be complemented with domain-specific metrics for scientific tasks. Unlike previous work, we implement instance-specific evaluation grounded in expert preferences, providing LLMs with detailed guidance on how each aspect should be addressed. With similar motivation, \citet{Chakrabarty:2025} train reward models for writing quality assessment, mainly focusing on creative writing such as literary fiction and marketing.

To sum up, to the best of our knowledge, our work is the first of its kind to 1) conceptualize and develop automated RW evaluation as an expert domain task with domain-specific utilities, and 2) develop text generation evaluation techniques beyond LLM-as-a-judge systems that can effectively address their limitations.

\section{Methodology}

\subsection{Dataset}

Previous studies focusing on RW section or citation text generation have utilized abstracts, metadata, citation intent or example citation sentences as the primary sources of context for cited papers \cite{Li:2024}. However, these materials fall short of providing sufficient information to disclose the relations between papers. 
On the other hand, introduction sections contain core essential information such as addressed problem, employed methodology, contributions, and results of the papers with minimal addition to the context length. Therefore, we complement abstracts with introduction sections to provide richer context for generation\footnote{Full-text usage would be ideal, but for papers with many cited references it can be impractical due to context-length limitations and evaluation cost.}.

For citing (main) papers, we use the open-license subset of the unarXive~\cite{Saier:2023} as main source and complement with OARelatedWork\footnote{Despite claiming full cited paper text coverage, only a limited subset does; manual curation is still needed.} \cite{Docekal:2026} (\textit{title, abstract, introduction, and related work}). We selected Natural Language Processing (NLP) and Computer Vision (CV) papers published in top-tier venues to (1) increase data quality, (2) maintain the feasibility of subsequent expert study, which is a common practice among previous work \cite{Xing:2020,Ge:2021,Li:2022}, and (3) ensure broader coverage and improve the generalizability. To ensure data quality, several stages of the pipeline, especially aggregating a complete set of cited papers from heterogeneous sources, are carefully curated by manual effort. The final dataset contains 50 main papers with an average of 13.50 cited papers each, totaling 725 papers. Further details of dataset construction are provided in Appendix \ref{app:data}.

\subsection{Evaluation criteria}\label{sec:eval}

We define a set of hard and soft constraints~\citep{Dutta:2025} for evaluating the generated RW sections based on previous theoretical work focusing on \textit{how to write a good related work section or conduct a literature review}~\cite{Randolph:2009,Jaidka:2013,Teevan:2023}. Hard constraints represent the essential requirements that the generated text must satisfy to be qualified as a valid RW section. Soft constraints define the grounds for an author's preferences among multiple valid drafts. We use gold RW sections as a proxy for author preferences for soft constraint evaluation. We note that gold RW sections serve as observable examples of a specific expert’s stylistic and structural preferences within a given context, not a reference for the uniquely optimal output. Alternative but equally valid outputs can still be integrated to \framework\ by modifying the soft constraint specifications. The hard constraints are as follows:

\paragraph{Citation Verification:} To verify the citations, we compute the fraction of papers (from the provided list of papers) \textit{not cited} in the generated RW as Missing Ratio and the fraction of cited ones not in the original list as Hallucination Ratio.

\textbf{Faithfulness}: We define faithfulness as the factual alignment between each generated citation sentence and the content of the corresponding cited paper provided to the generator (title, abstract, and introduction). A citation is faithful only if all claims are fully supported by the source; any unsupported or contradictory statements constitute a violation. Evaluating against the same context given to the generator ensures that models are not penalized for information they did not receive. Since factual errors in RW sections undermine reliability, we treat faithfulness as a hard constraint. We formulate this as an NLI (natural language inference) problem, following previous works that have adopted similar approaches focusing on summarization \cite{Scire:2024}, factual consistency \cite{Zha:2023, Honovich:2022}, and text generation with citations \cite{Gao:2023}. Details of the faithfulness ratio calculation are given in Appendix \ref{app:coherence}.

\paragraph{Positioning Existence:} One of the essential functions of the RW sections is to differentiate the presented work from previous studies. We use the term ``positioning'' to refer to how authors situate their work within the existing research landscape, highlighting contributions that distinguish it from others as a signal of novelty. Since RW sections should not be pure summaries of previous works, we evaluate whether generated RW sections include statements highlighting the positioning of the main paper in the literature. Further details and example statements can be found in Appendix \ref{app:positioning}.

Since expert preferences in scientific writing encapsulate numerous features, it is not feasible to incorporate every one of them. Therefore, we design our soft constraints to focus on aspects that are (1) explicitly specifiable, (2) measurable, and (3) controllable within an automated evaluation loop. These criteria cover both structural features and how experts contextualize previous work when framing their contributions. In short, our design represents a tractable step toward modeling expert-informed evaluation criteria. The following are the soft constraints we consider:

\paragraph{Length:} Depending on the type of a paper (e.g., long/short research papers, surveys) and the author preferences, the length of RW sections varies. We check whether the number of tokens in the generated RW section falls in an interval within a tolerance ratio around the number of tokens in the gold RW sections. Details of tolerance ratio and the length evaluation are provided in Appendix \ref{app:length}.

\paragraph{Citation Emphasis:} In RW sections, some papers are discussed in detail, while others are briefly mentioned and included in group citations. We measure how much content is allocated for each citation. For each citation, we define the allocated content as the sentences including the corresponding citations and the follow-up sentences that do not contain any other citation and do not start a new paragraph. We calculate the ratio between the number of tokens in the allocated content and the total number of tokens in the generated RW section. Then, we compare this ratio for the generated draft and the gold RW section. Similar to the length constraint, we check whether the emphasis score for the generated draft is within the desired interval constructed by gold paper values with a tolerance ratio. Finally, we average individual citation emphasis values to get an overall score for a generated RW section. The process is explained algorithmically in Appendix \ref{app:citation_emp}.

\paragraph{Positioning Type:} Similar to other soft constraints, the expression of contribution and positioning of the paper depends on the author's writing preferences. We consider two types of expressions: (1) the contribution and the position of the paper are provided in each paragraph in accordance with the corresponding subject matter of the paragraph, (2) the contribution and the position of the paper are emphasized in the final paragraph by addressing the points mentioned in all previous paragraphs. We use a joint prompting strategy, detecting both the existence and type of an expression. If it exists, we check that the predicted type is the same as the type specified in the prompt during generation.        

\paragraph{Positioning Ratio:} It is possible that individual paragraphs may partially satisfy the expected type of expression. If the positioning type is each paragraph emphasis, we check whether each paragraph includes a contribution expression. For the other, we check whether the final paragraph addresses the points of each earlier paragraph while emphasizing the contribution or positioning. Then, we calculate the ratio of positively evaluated paragraphs.

\subsection{Evaluation framework}

Faithfulness and positioning related criteria require natural language understanding. Language models are a natural choice in such cases. However, our preliminary experiments show that applying vanilla zero-shot LLM-as-a-judge remains insufficient for expert domain evaluations. We identify the main reason as the absence of context information indicating a specific evaluation criterion and what it means to satisfy (or not) that\footnote{While few-shot examples are known to improve performance, they cannot be incorporated when using an end-to-end judge due to context-length constraints.}. For each possible outcome of a specific evaluation, we include an example along with a reasoning component that explains the expected outcome. Since finding failing examples for specific aspects is non-trivial, we generate synthetic examples using LLMs prompted to make deliberate mistakes (authors manually check these instances). We present our examples in Appendix \ref{app:contrastive_few_shot} for each LLM-based evaluation.

\framework\ employs an iterative algorithm where generation and evaluation are interleaved, simulating multi-turn human-AI interaction. Henceforth, we call the LLM under evaluation as \textit{generator}. Given the details (title, abstract, and introduction) of the main and cited papers and the task prompt, the generator comes up with a draft that is evaluated against the adopted criteria. Evaluation scores and justifications are aggregated into an evaluation report, which the generator model itself then converts into proxy natural language feedback (prompt in Table~\ref{tab:feedback_prompt}). We manually inspected the generated feedback to ensure that it faithfully reflects the report items. This feedback (a sample is provided in Appendix~\ref{app:framework}) guides the generation of the next draft to better align with expert preferences.

\section{Experiments}

\begin{table}[]
\centering
\small
\begin{tabular}{c|ccc} 
\hline
\textbf{Model} & \textbf{Faithfulness} & \textbf{Pos. Type} & \textbf{Pos. Ratio} \\
\hline
GPT-4o & \textbf{0.82} & 0.94 & \underline{0.92} \\
o3-mini & 0.70 & \textbf{1.00} & \textbf{1.00} \\
Llama 3.3 & 0.72 & 0.92 & \textbf{1.00} \\
Gemma 3 & \underline{0.80} & \underline{0.96} & 0.88 \\
\hline
\end{tabular}
\caption{Accuracy of preliminary evaluations. The best results for corresponding task are in bold. Positioning existence is jointly implemented with positioning type.}
\label{tab:pre-eval}
\end{table}

\paragraph{Selecting evaluator LLMs.} Toward implementing LLM-based evaluation of faithfulness and positioning, we experiment with four widely used LLMs: GPT-4o (2024-11-20) \cite{Openai:2024}, o3-mini (2025-01-31) \cite{Openai:2025}, Gemma 3 (27b) \cite{Gemmateam:2025}, and Llama 3.3 Instruct (70b) \cite{Grattafiori:2024}. We create meta-evaluation benchmarks consisting of 50 samples for each criterion: faithfulness, positioning type, and positioning ratio. To make each benchmark balanced, we synthetically generate data instances by mismatching cited papers and citation sentences for faithfulness evaluation and rewriting RW sections in our dataset according to specific positioning styles (per-paragraph positioning, aggregate positioning, no positioning) via GPT-4o. Since labels follow directly from how each instance is constructed, the authors manually verified that each generated instance matches its intended label. Evaluations are repeated three times with a temperature of $0.8$, and the final decision is made by majority voting to increase robustness. We provide the prompts in Appendix \ref{app:prompts_eval}. We report the preliminary results in Table \ref{tab:pre-eval}, indicating a clear gap between the proprietary and open models. Subsequently, in Precise\framework, we use GPT-4o and o3-mini for faithfulness and positioning evaluations, respectively. In Open\framework, we use Gemma 3 for faithfulness and positioning type, while Llama 3.3 for positioning ratio.

\paragraph{Domain expert evaluation.}
After model selection, we conduct an expert evaluation study to validate \framework's judgments. Human experts interact with a pair of generator models simultaneously. Both models use the same main paper and cited papers to generate RW sections. At each iteration, the experts evaluate the generated drafts in terms of faithfulness, positioning, and feedback (instruction) following capabilities, and provide  independent feedback to each model. Since the objective of \framework\ is not to measure absolute conformity to a single RW section, but to evaluate controllability and responsiveness to expert feedback, no reference gold RW sections are provided. Each expert shapes their own RW section independently. Since our dataset includes multiple domains, we conducted expert evaluation separately across domains. This design choice helps to validate the generalization capabilities of \framework. 16 senior PhD students and postdoctoral researchers (10 NLP, 6 CV), with 10.9 avg. published papers, participated in our study.

As baselines, we select three open models with sufficient context length, Self-Taught Evaluator (STE)\footnote{\href{https://huggingface.co/facebook/Self-taught-evaluator-llama3.1-70B}{Self-taught-evaluator-llama3.1-70B}}, DeepSeek-GRM\footnote{\href{https://huggingface.co/BBQGOD/DeepSeek-GRM-16B}{DeepSeek-GRM-16B}}, and Nemotron\footnote{\href{https://huggingface.co/nvidia/Llama-3_3-Nemotron-Super-49B-GenRM}{Llama-3\_3-Nemotron-Super-49B-GenRM}} representing strong recent reward models adopted in LLM-as-a-Judge systems \cite{Wang:2024,Wang:2024self,Chen:2025,Liu:2025dsgrm,Whitehouse:2025,Saha:2025}. We compute (for each criterion) the fraction of matching judgments between the expert and an evaluation framework (two variants of \framework\ and baselines). We provide further implementation details in Appendix \ref{app:human_eval}.

\paragraph{RW Generator evaluation.}
After validating \framework\ in expert evaluation, we finally use it in the RW generation pipeline for five iterations. We first evaluate 10 LLMs of varying scales and families on a subset of our data as generators by Open\framework: GPT-4o-mini, GPT-4o, o3-mini, Gemma 3 (27b), Llama 3.3 Instruct (70b),
Deepseek-R1 (70b)~\cite{Deepseekai:2025},
Mistral (7b)~\cite{Jiang:2023},
Phi-4 (14b)~\cite{Abdin:2024},
Qwen 2.5 (72b)~\cite{Qwen2.5:2025},
and Qwen 3 (30b)~\cite{Yang:2025}. 
To minimize costs for proprietary evaluators, four of these models (selected via systematic sampling from all models, ranked by average scores) are evaluated using Precise\framework: o3-mini, GPT-4o, Llama 3.3, and Gemma 3. Further details and prompts are given in Appendix \ref{app:prompt_gen} and \ref{app:exp_pipeline}.

\begin{figure*}[!t]
\centering
\includegraphics[width=1\linewidth]{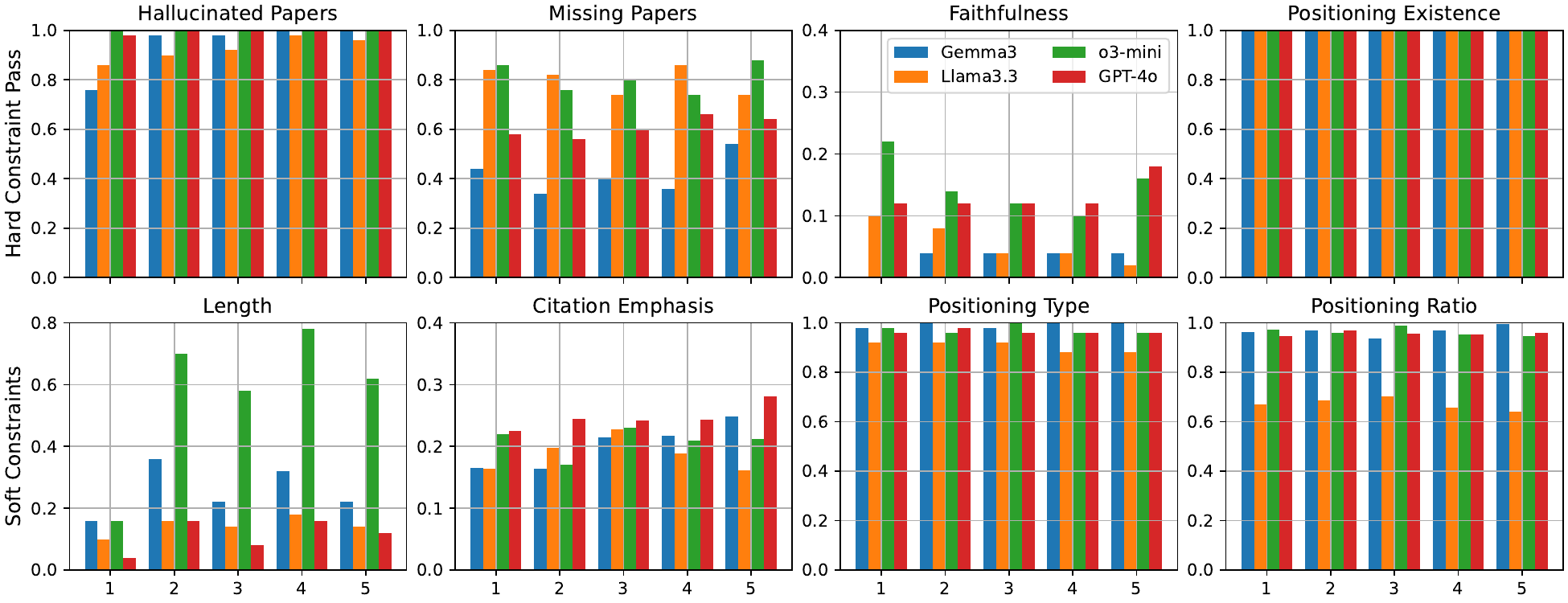}
\caption{Overall Precise\framework\ results with four generator LLMs. Scores for each criterion are averaged across RW sections. Faithfulness is the hardest test to pass, while all models deliver perfect scores for Positioning Existence.}
\label{fig:full_pipeline}
\end{figure*}

\section{Results}\label{sec:results}

\begin{table}[!t]
\centering
\resizebox{\columnwidth}{!}{
\begin{tabular}{l|ccc|ccc}
\hline
\multirow{2}{*}{\textbf{Framework}} & \multicolumn{3}{c|}{\textbf{NLP Domain}} & \multicolumn{3}{c}{\textbf{CV Domain}} \\
\textbf{} & \textbf{F} & \textbf{P} & \textbf{Fb} & \textbf{F} & \textbf{P} & \textbf{Fb} \\
\hline
STE & 0.53 & 0.63 & 0.47 & 0.44 & 0.39 & 0.39 \\
DS-GRM & \underline{0.66} & 0.63 & 0.50 & \underline{0.67} & 0.50 & \textbf{0.67} \\
Nemotron & 0.41 & 0.53 & 0.28 & 0.44 & 0.28 & 0.44\\
\hline
Open\framework & \underline{0.66} & \underline{0.66} & \underline{0.66} & \underline{0.67} & \textbf{0.72} & \textbf{0.67} \\
Precise\framework & \textbf{0.78} & \textbf{0.75} & \textbf{0.69} & \textbf{0.72} & \textbf{0.72} & \textbf{0.67} \\
\hline
\end{tabular}
}
\caption{Match rate with expert judgments. F: Faithfulness, P: Positioning, Fb: Feedback. We report confidence intervals in Table \ref{tab:confidence} in Appendix \ref{app:human_eval}.}
\label{tab:human_eval}
\end{table}

\paragraph{Expert evaluation across domains.} Table \ref{tab:human_eval} shows the accuracy of \framework\ variants and baselines across domains. In NLP (primary) domain, we observe that both variants of \framework\ consistently outperform judge models. The weak performance of the baselines to detect the presence (or lack) of faithfulness indicates the lack of domain-specific deep reasoning ability in specialized judge models. Evaluating the general feedback following capabilities is more challenging than well-defined, decomposed evaluation aspects, possibly due to the lack of context of human cognitive factors. While the Open\framework\ lags behind the proprietary one, it is still moderately aligned with expert judgment. Though \framework\ is designed to deliver cardinal scores, it closely matches the ordinal expert judgment, implying the robustness of \framework\ as an evaluation framework.

To validate domain generalization, we repeated the same experiment with CV experts and papers. CV domain results are consistent with the NLP domain, as Open\framework\ and Precise\framework\ continue to outperform the baselines. Furthermore, we observe that baseline models fluctuate considerably, while \framework\ is more robust across domain changes. This shows that \framework\ is a reliable framework generalizable across varying domains.

Upon validating \framework's evaluation robustness via experts across different domains, we now test how existing LLMs fare in different hard and soft constraint satisfaction for RW generation, and how well they adapt to dynamic user preferences. As \framework\ is reliable for both domains, we present results jointly. Due to space constraints, results from Open\framework\ are in Appendix \ref{app:exp_open_eval}.

\paragraph{Hard constraint satisfaction.}
Figure \ref{fig:full_pipeline} summarizes the results for different evaluation criteria across iterations. Three overall observations can be made: 1) for very few papers, all the hard constraints are met in the first iteration, signifying that \textit{even the best current models lack the ability to reason and write a valid RW section on their own}, 2) citation faithfulness is the hardest test to pass, i.e., \textit{LLMs are limited in their ability to deeply reason with scientific papers} and 3) central to human-AI collaboration, \textit{using feedback to improve hard constraint satisfaction is generally lacking} and varies from model to model (see Appendix \ref{app:exp_delta} for improvement trends of each model). Within the scope of our dataset, o3-mini is generally best in terms of passing the hard constraints\footnote{Possibly due to its STEM-focused training as a reasoning model: https://openai.com/index/openai-o3-mini/ }: no imaginary citation, no RW section without positioning statements, and not missing any papers to cite in more than 70\% cases. In the first iteration, o3-mini fares in the faithfulness test substantially better than other models (all accurate citations in 20\% of the generated drafts as opposed to 10\% by Llama-3.3). This difference quickly diminishes with feedback: while rest of the models do not improve, o3-mini starts failing more frequently. Feedback is most helpful for correcting hallucinated citations. GPT-4o generally improves better than other models with feedback, across all four criteria. Across all four models, failing to cite the provided papers is more common than hallucinating non-existent ones. Similar patterns are evident in the evaluation of Open\framework: while Deepseek-R1 and GPT-4o-mini are great at generating faithful citations, they fail to cite all provided papers a significant amount of time; GPT-4o, Qwen 3 and 2.5 demonstrate the exact opposite behavior.

\begin{figure*}[!t]
\centering
\includegraphics[width=1\linewidth]{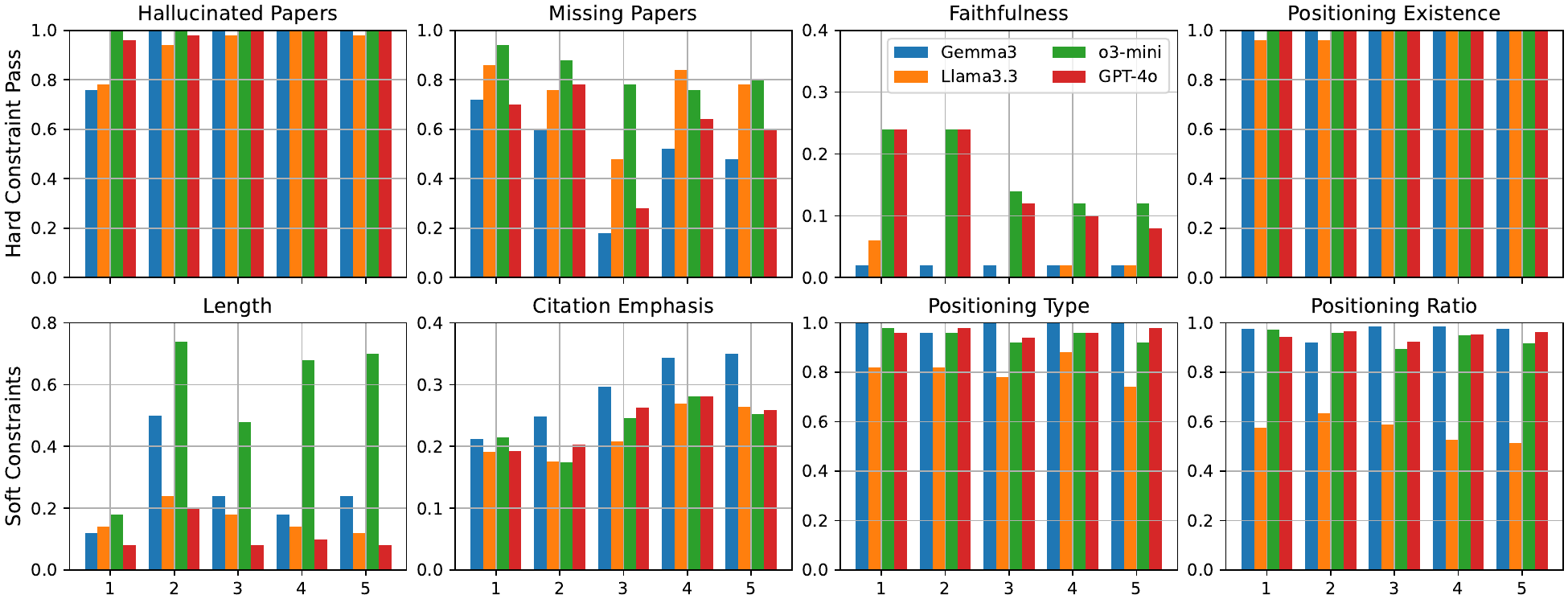}
\caption{Adaptability to new paper introduction evaluated by Precise\framework. Missing paper increases at the point of new paper introduction (3rd iteration), implying the inability to accommodate new information.}
\label{fig:new_paper}
\end{figure*}

\begin{figure*}[!t]
\centering
\includegraphics[width=1\linewidth]{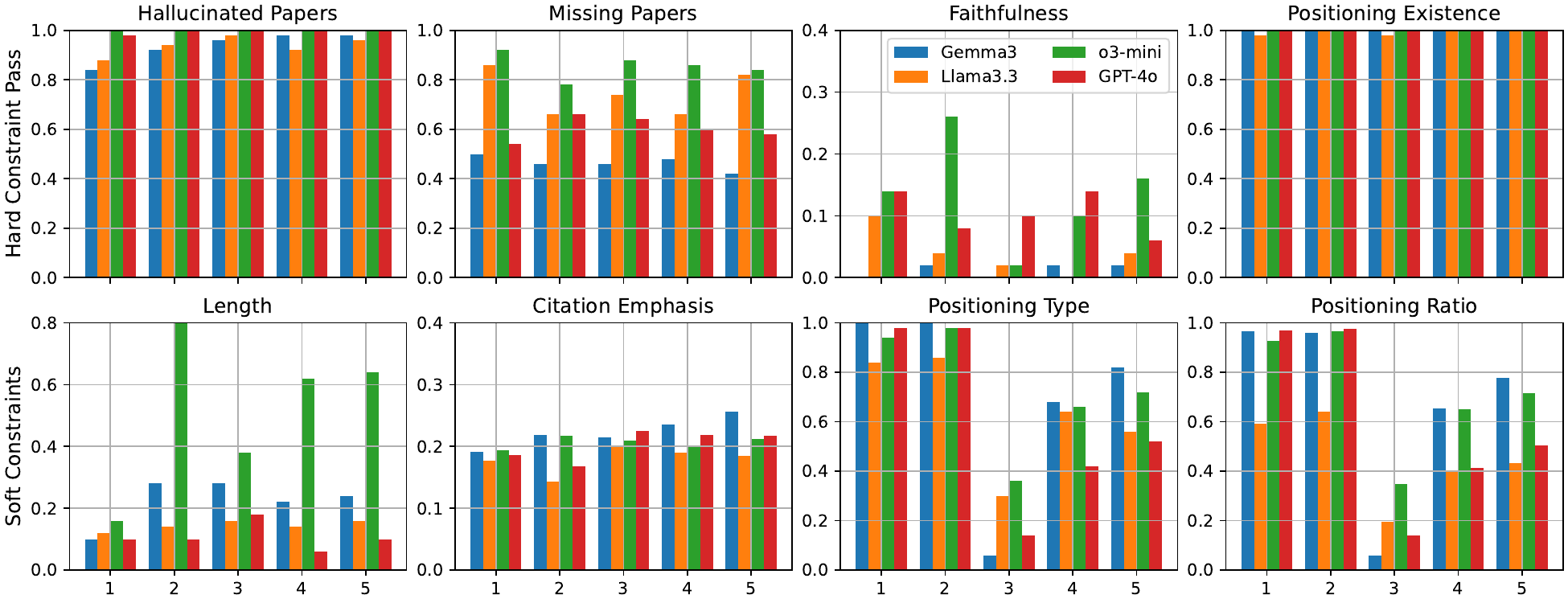}
\caption{Adaptability to style change evaluated by Precise\framework. Positioning type and ratio-based score drops at the point of change (3rd iteration), and models struggle to acquire original performance even after repeated feedback.}
\label{fig:style_change}
\end{figure*}

\paragraph{Soft constraint performance.} 
Due to the overgeneration tendency of LLMs~\cite{Singhal:2024}, lengths of the generated RW sections typically overshoot in the first iteration. o3-mini emerges as the best model to follow the feedback and adjust the length accordingly. \textit{The rest of the models struggle to revise the generated draft's length according to explicit instructions}. Gemma 3 stands out for consistent improvement across iterations for citation emphasis. However, there is a large gap in incorporating author preferences to adjust allocated citation content across all the models. Similar to the positioning existence, all models almost perfectly reflect the expected positioning type consistently. This pattern carries over to the individual evaluation of the paragraphs, except Llama 3.3. 

\paragraph{Adaptability to new paper introduction.}
We investigate the effects of adding new papers \textit{during} the interaction to simulate a realistic human-AI interaction. We start the generation without providing 25\% (remainders rounded) of the cited papers. Then, we introduce the held-out papers at the start of the third iteration. Results are presented in Figure \ref{fig:new_paper}. Failing due to missing citations peaks at the third iteration, implying that \textit{models cannot integrate the new content mid-interaction properly, except for o3-mini}. With feedback, all models bounce back. Interestingly, dynamically introducing papers helps all models to satisfy citation emphasis constraint better than the static variant. The increasing trend, particularly with Gemma 3, indicates that \textit{the gradual introduction of papers can facilitate better emphasis alignment}. The remaining evaluation aspects mostly stay the same.  

\paragraph{Dynamic style change request.}
In this setup, we change the expected positioning expression types starting from the third iteration. The evaluation results are provided in Figure \ref{fig:style_change}. Similar to experiments with introduction of new papers, we observe that the LLMs cannot immediately adapt to the style changes of the authors: positioning type and positioning ratio show a substantial decline after the third iteration. Although performance increases with feedback, \textit{two iterations after style changes seem not sufficient to restore the initial performance}. Furthermore, this setup also shows that our evaluation schemes are capable of detecting LLM failures against changing user preferences in a realistic simulation of human-AI interaction for an expert domain task. We provide full numeric results and cross-paper variability along with additional details on our experimental settings in Appendix \ref{app:full_results}.

\paragraph{Error Analysis.} To complement our evaluations, we analyze the errors made by generator models and discuss potential solutions. We first observe that LLMs can struggle to follow specified citation rules. Incorrect citation formats (e.g., author-year instead of numeric) cause parsing errors in citation sentences, which negatively affect missing paper and faithfulness measurements. In the feedback setting, as the number of items requiring correction increases, LLMs may overlook some items, or modifying one feature can sometimes deteriorate another one. This pattern has also been reported by \citet{Dutta:2025}. One solution can be dividing feedback into smaller actions and applying step-by-step updates. However, this approach introduces a trade-off between the quality of generated content and the increased number of LLM calls, resulting in additional financial and computational costs.

\section{Conclusion}

In this work, we introduce \framework, a comprehensive evaluation framework for RW generation, designed to bridge current gaps in evaluating automated solutions in expert domains. It consists of multiple modules, each specialized in a different aspect of the task based on expert preferences, providing greater granularity in interpreting evaluation and improving subsequent generations. Demonstrating robust generalization across domains, \framework\ offers a reliable evaluation framework applicable to diverse research communities beyond NLP, available in both a high-precision proprietary and a cost-efficient open-source variant. \framework\ simulates human-AI collaboration in scientific writing with dynamically evolving human preferences. The outputs of the modules can serve as faithful proxies for human judgment in assessing LLM performance, reducing the cost of human-in-the-loop experimentation.

\section*{Limitations}

While LLM-driven evaluations in \framework\ are susceptible to errors due to a lack of domain-grounded reasoning, this is a known challenge across LLM-based evaluation frameworks and motivates our use of multiple specialized modules and contrastive few-shot examples to mitigate potential errors. Our framework also shows strong alignment with human expert judgments and consistent performance across both Open\framework\ and Precise\framework\ variants provide empirical support for its reliability. 

We prioritize dataset quality and evaluation diversity over scale, as very large datasets incur significant LLM call costs (Appendix \ref{app:full_results}) and manual curation overhead. Our dataset is carefully constructed to provide sufficient instances for observing LLM capabilities on RW generation while comprising multiple domains. Immediate future work can involve incorporating search agents that look for relevant papers and evaluate their combined performance. Furthermore, nuanced, stylistic author preferences (e.g., active vs passive voice, stressing certain concepts, etc.) can be explored. 

Finally, the generator and evaluator models were fixed at the time of our experiments; given the pace of LLM development, newer models may behave differently. Since \framework\ is model-agnostic, it can be applied directly to newer generators and judges.

\section*{Acknowledgments}

This work was funded by the ``Modeling Task-oriented Dialogues Grounded in Scientific Literature" project in partnership with Amazon Alexa. We gratefully acknowledge the support of Microsoft with a grant for access to OpenAI GPT models via the Azure cloud (Accelerate Foundation Model Academic Research). Furkan \c{S}ahinu\c{c} is supported by the Konrad Zuse School of Excellence in Learning and Intelligent Systems (\href{https://eliza.school/}{ELIZA}) through the DAAD programme Konrad Zuse Schools of Excellence in Artificial Intelligence, sponsored by the Federal Ministry of Education and Research.

\bibliography{camera_ready}

\appendix

\section{Dataset}\label{app:data}

To generate high quality RW sections, the models should have access to a complete set of cited paper information. However, content retrieval for cited papers is a remarkably challenging task because all cited papers cannot be accessed from a single common source. We first start with the S2ORC \cite{Lo:2020} and OARelatedWork \cite{Docekal:2026} dataset that provides the required content for 57\% of the cited papers. For the rest, content is collected from the PDFs (retrieved via URLs from metadata) using S2ORC parser tool\footnote{https://github.com/allenai/s2orc-doc2json}. Any parsing problems are corrected manually. In addition, we exclude any cited paper that lacks open-license from the related work sections. Text segments associated with the removed citations are also deleted. If the removed citations are critical for the related work section or the remaining content after removal became too short, we drop the citing paper altogether from the dataset. 

\section{Evaluation Methodology}

\begin{algorithm}[ht]
\small
\caption{Citation Emphasis Evaluation}
\label{alg:citation_emp}
$R^{Gold}$: Gold RW section \\
$R^{Gen}$: Generated RW section \\
$t$: Tolerance threshold ratio 
\begin{algorithmic}[1] 
\State $eval = []$
\State $emp = \{0,0,..0\}$
\For{$paragraph \in R^{Gen}$}
\State $currentIds = []$
\For{$sentence \in paragraph$}
\State $citedIds = extractCitation(sentence)$
\If {$citedIds \neq \emptyset$}
\State // New citations in the sentence
\State $currentIds = citedIds$
\EndIf
\If{$currentIds \neq \emptyset$}
\State // No paragraph start
\State $emp[currentIds] \mathrel{+}= \frac{Token(sentence)}{TotalToken}$
\EndIf
\EndFor
\EndFor
\For{$citedId \in R^{gold}$}
\State $upper = (1+t)*emp^{gold}[id]$
\State $lower = (1-t)*emp^{gold}[id]$
\If {$emp[citedId] \in [lower, upper]$}
\State $eval[citedId] = 1$
\Else
\State $eval[citedId] = 0$
\EndIf
\EndFor
\State \textbf{return} $mean(eval)$
\end{algorithmic}
\end{algorithm}

\subsection{Faithfulness}\label{app:coherence}

We consider each citation element while checking the faithfulness of the citation sentences. We use the abstract and introduction sections of the cited papers as reference points to be compared with the citation sentences. If there are multiple citations in the sentence, we do not evaluate all citations at once. Instead, a separate evaluation is performed for each cited paper using its specific citation number. Finally, we calculate the positive outcome ratio in all evaluated sentence-paper pairs. A valid RW section should have a perfect (i.e., 1.0) score. Therefore, we report the average ratio of outcomes that are equal to $1.0$ (i.e., passing the faithfulness hard constraint).

\subsection{Positioning}\label{app:positioning}

Although there is no single way to position a paper in the literature, there are some common patterns used in RW sections to clearly emphasize the differences of the work, such as "\textit{Although prior research does [X], our work addresses [Y], providing [new insight / practical solution / improvement]}." or "\textit{In this work, we fill the gap [X] by doing [Y, Z], unlike previous work}." While our contrastive few-shot examples implicitly incorporate such patterns, we avoid providing explicit instructions that enforce a specific template to minimize potential bias.

\subsection{Length}\label{app:length}

\begin{figure}
\centering
\includegraphics[width=\linewidth]{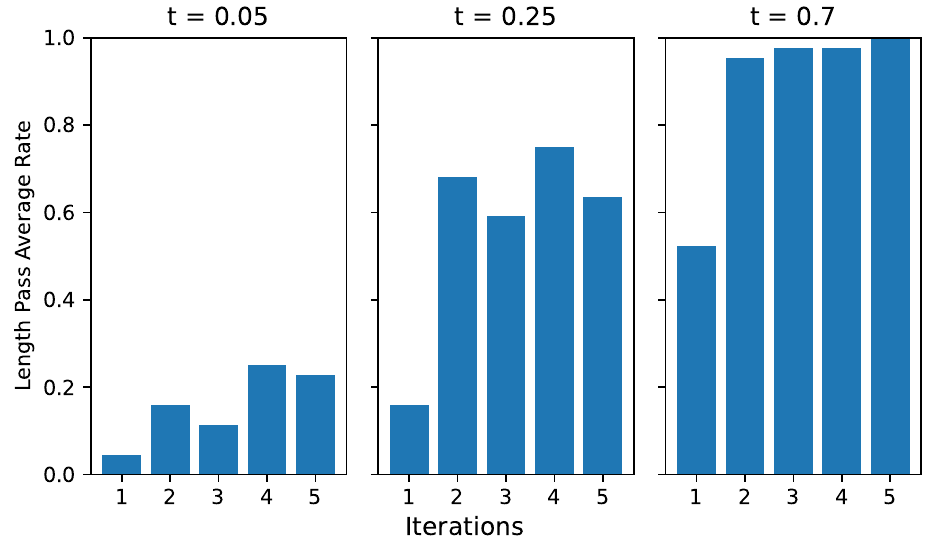}
\caption{Length constraint average pass rate for generated RW sections in different iterations with varying tolerance ratio values. Low and high tolerance values lead to converge both ends, respectively, preventing to observe the effect of provided feedback across iterations.}
\label{fig:tolerance}
\end{figure}

The length evaluation function is given as follows.

\begin{equation*}
f_L(x) =
\begin{cases}
1 & \text{if } x \in [(1-t)*T, (1+t)*T] \\
0 & \text{otherwise}
\end{cases}
\end{equation*}

where $t \in (0,1)$ is the tolerance ratio and $T$ is the number of tokens in gold related work section. In Figure \ref{fig:tolerance}, we experiment with different tolerance ratio values with the RW sections generated by o3-mini. As expected, small tolerance ratios cause oscillations between too long and too short RW sections, resulting in very low length pass rates. In contrast, large tolerance ratios pass the length constraint most of the time due to loose control mechanism, making it difficult to assess whether LLMs were actually generating sections of similar length to the reference. Values around $0.25$ allow us to detect meaningful deviations across iterations and between different LLMs. Therefore, $0.25$ tolerance ratio is used in the main experiments.

\subsection{Citation Emphasis}\label{app:citation_emp}

We provide an algorithmic representation of citation emphasis evaluation in Algorithm \ref{alg:citation_emp}. Following the same reasoning as the length criteria, the tolerance ratio is set to $0.25$.

\subsection{Contrastive Few-shot Examples}\label{app:contrastive_few_shot}

We provide contrastive few-shot examples that we use in our evaluation setup in Tables \ref{tab:coherence_example_1} and \ref{tab:coherence_example_2} for faithfulness evaluation, Tables \ref{tab:cont_type_example_1}, \ref{tab:cont_type_example_2}, and \ref{tab:cont_type_example_3} for positioning type evaluation, Tables \ref{tab:cont_direct_eval_examples} and \ref{tab:cont_pairwise_eval_examples} for positioning ratio evaluation.

\begin{algorithm}
\small
\caption{Pipeline}
\label{alg:pipeline}
\textbf{Dataset}: $D=\{( C_i, \{R_{i,j}\}_{j=1}^{n_i}, y_i)\}_{i=1}^{N}$\\
// $C_i=(T_i^c, A_i^c, I_i^c)$: Citing paper $i$ \\ 
// $R_{i,j}=(T_{i,j}^r, A_{i,j}^r, I_{i,j}^r)$: Cited papers in $C_i$\\
// $y_i$: Related work section of the citing paper $C_i$ \\
// $T$: Title, $A$: Abstract, $I$: Intro.
\begin{algorithmic}[1] 
\For{$i \in {1,..,N}$}
\For{$k \in {1,..,K}$}
\If {$k == 1$}
\State $\hat{y}_i^k$ = $\text{genDraft}(x, C_i, \{R_{i,j}\}_{j=1}^{n_i})$
\Else
\State $\hat{y}_i^k$ =$\text{genDraft}(x, \hat{y}_i^{k-1}, f^{k-1}, C_i, \{R_{i,j}\}_{j=1}^{n_i})$
\EndIf
\For{\textbf{each} evalModule $m \in M$}
\State $e_m^k, j_m^k$ = $\text{evalModule}_m(\hat{y}_i^k)$
\EndFor
\State $e_M^k, j_M^k$ = $\text{aggregate}(\{e_m^k\}_{m \in M}, \{j_m^k\}_{m \in M})$
\State $f^k$ = $\text{genFeedback}(e_M^k, j_M^k)$
\EndFor
\EndFor
\State \textbf{return} $\{\hat{y}_i^K\}_{i=1}^{N}$
\end{algorithmic}
\end{algorithm}

\subsection{Evaluation Framework}\label{app:framework}

We provide the algorithmic representation of \framework\ in Algorithm \ref{alg:pipeline}. An example of feedback generated by o3-mini in our experiments is as follows: "\textit{Reduce the overall length by about 25\% while rebalancing citation emphasis—dial back the extensive focus on paper 5 and provide more detail on paper 6. Revise the problematic sentence regarding paper 4 to ensure that the cited work's context aligns with the claim verification framework described. Finally, better integrate the concrete methods (like multi-task feature filtering and social dynamics insights) from earlier paragraphs into the final summary of contributions so that the introduced unified dual-view framework is clearly connected to the compared approaches.}"

\section{Prompts}

\subsection{Evaluation Prompts}\label{app:prompts_eval}

We present the prompts we used in the evaluation stages of \framework\ in Tables \ref{tab:coherence_prompt} for the faithfulness evaluation, \ref{tab:contribution_type_prompt} for the positioning type, and \ref{tab:cont_direct_eval_prompt} - \ref{tab:cont_pairwise_eval_prompt} for the positioning ratio.

\subsection{Generation Prompts}\label{app:prompt_gen}

We present the prompts we used in the generation stages of \framework\ in Tables \ref{tab:first_draft_prompt} and \ref{tab:next_draft_prompt} for draft generation, Table \ref{tab:feedback_prompt} for feedback generation.

\section{Experimental Details}

\subsection{Pipeline Configurations}\label{app:exp_pipeline}

We use the vLLM framework\footnote{https://docs.vllm.ai/en/stable/} to run open-weight LLMs locally with 4-bit quantization on a single NVIDIA A100 GPU with 80GB memory. For the OpenAI models, we used API version \verb|2025-03-01-preview|. We set the temperature value to 0.8 for all models, except o3-mini, which does not support temperature adjustment. The remaining model parameters were left at their respective default values.

We use the structured output feature of the vLLM and API libraries to facilitate parsing of LLM outputs for evaluation. We leverage JSON schema as the response format where reasoning and evaluation verdict are two output components. To parse citation sentences, we utilize \verb|en_core_sci_sm| model from the ScispaCy library\footnote{https://allenai.github.io/scispacy/}.

\begin{figure*}
\centering
\includegraphics[width=\linewidth]{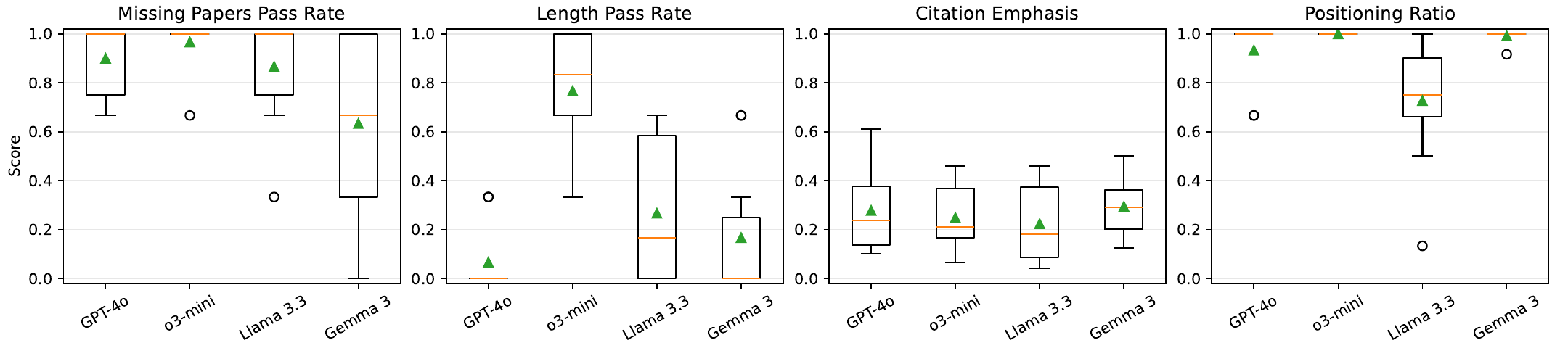}
\caption{Across-paper score distributions of four generators at the final iteration (10-instance subset; per-paper scores averaged over three runs). Boxes show the interquartile range, orange lines the median, green triangles the mean, and circles outliers.}
\label{fig:across_paper}
\end{figure*}

\subsection{Open Evaluators}\label{app:exp_open_eval}

We report our evaluation results obtained with Open\framework \ in Table \ref{tab:open_eval_hard} for hard constraints and Table \ref{tab:open_eval_soft} for soft constraints. Scores are averages across papers. We report full pipeline results without any new paper introduction or style changes. For Precise\framework \ experiments we sampled four models to create a comprehensive benchmark that represents a range of different performances across different model families, such as reasoning and instruction-tuned models, within cost constraints.

\begin{table*}[ht]
\centering
\setlength{\tabcolsep}{1.2mm}
\resizebox{\textwidth}{!}{
\begin{tabular}{l|ccccc|ccccc|ccccc|ccccc}
\hline
\textbf{Hard Const.} & \multicolumn{5}{|c}{\textbf{Hallucinated Papers}} &  \multicolumn{5}{|c}{\textbf{Missing Papers}} &  \multicolumn{5}{|c}{\textbf{Faithfulness}} & \multicolumn{5}{|c}{\textbf{Positioning Existence}} \\
\hline
Deepseek-R1 & 1.0 & 1.0 & 1.0 & 1.0 & 1.0 & 
0.3 & 0.7 & 0.5 & 0.9 & 0.7 & 
0.6 & 0.8 & 0.7 & 0.8 & 0.8 & 
1.0 & 1.0 & 1.0 & 1.0 & 1.0 \\

Gemma 3 & 0.8 & 1.0 & 1.0 & 1.0 &  1.0 &  
0.7 & 0.5 & 0.7 & 0.8 & 0.6 & 
0.1 & 0.1 & 0.3 & 0.3 & 0.5 & 
1.0 & 1.0 & 1.0 & 1.0 & 1.0 \\

GPT-4o-mini & 1.0 & 1.0 & 1.0 & 1.0 & 1.0 &
0.5 & 0.5 & 0.8 & 0.7 & 0.6 &
0.9 & 0.8 & 0.7 & 0.6 & 0.7 & 
1.0 & 1.0 & 1.0 & 1.0 & 1.0 \\

GPT-4o & 1.0 & 1.0 & 1.0 & 1.0 & 1.0 &
0.9 & 0.9 & 1.0 & 1.0 & 1.0 &
0.4 & 0.3 & 0.4 & 0.4 & 0.4 &
1.0 & 1.0 & 1.0 & 1.0 & 1.0 \\

Llama 3.3 & 0.7 & 0.8 & 0.8 & 0.9 & 0.9 &
1.0 & 0.9 & 0.9 & 1.0 & 0.9 &
0.1 & 0.1 & 0.0 & 0.0 & 0.0 & 
1.0 & 1.0 & 1.0 & 1.0 & 1.0 \\

Mistral & 0.8 & 0.7 & 0.9 & 0.9 & 0.9 &
0.3 & 0.5 & 0.2 & 0.3 & 0.3 &
0.1 & 0.1 & 0.1 & 0.1 & 0.1 & 
1.0 & 1.0 & 1.0 & 1.0 & 1.0 \\

o3-mini & 1.0 & 1.0 & 1.0 & 1.0 & 1.0 &
0.9 & 0.8 & 0.9 & 0.9 & 0.9 & 
0.4 & 0.4 & 0.5 & 0.5 & 0.4 & 
1.0 & 1.0 & 1.0 & 1.0 & 1.0 \\

Phi 4 & 1.0 & 1.0 & 1.0 & 1.0 & 1.0 &
0.4 & 0.5 & 0.6 & 0.6 & 0.6 & 
0.2 & 0.6 & 0.5 & 0.5 & 0.5 & 
1.0 & 1.0 & 1.0 & 1.0 & 1.0 \\

Qwen 3 & 1.0 & 1.0 & 1.0 & 1.0 & 1.0 & 
0.7 & 0.7 & 0.9 & 0.5 & 1.0 &  
0.1 & 0.2 & 0.1 & 0.2 & 0.0 &
1.0 & 1.0 & 1.0 & 1.0 & 1.0 \\

Qwen 2.5 & 1.0 & 1.0 & 1.0 & 1.0 & 1.0 &
0.5 & 0.6 & 0.5 & 0.4 & 0.7 &
0.2 & 0.4 & 0.2 & 0.3 & 0.3 & 
1.0 & 1.0 & 1.0 & 1.0 & 1.0 \\

\hline
\end{tabular}
}
\caption{Performance of different LLM generators in terms of the hard constraint passing rate, evaluated by Open\framework. While DeepSeek-R1 and GPT-4o-mini come as the best models in terms of citation faithfulness, they frequently fail to cite papers from the provided list.}
\label{tab:open_eval_hard}
\end{table*}

\begin{table*}[ht]
\centering
\setlength{\tabcolsep}{1.2mm}
\resizebox{\textwidth}{!}{
\begin{tabular}{l|ccccc|ccccc|ccccc|ccccc}
\hline
\textbf{Soft Const.} & \multicolumn{5}{|c}{\textbf{Length}} &  \multicolumn{5}{|c}{\textbf{Citation Emphasis}} &  \multicolumn{5}{|c}{\textbf{Positioning Type}} & \multicolumn{5}{|c}{\textbf{Positioning Ratio}} \\
\hline
Deepseek-R1 & 0.3 & 0.4 & 0.6 & 0.5 & 0.4 & 
0.16 & 0.27 & 0.22 & 0.25 & 0.27 & 
0.8 & 0.7 & 1.0 & 0.8 & 0.8 & 
0.59 & 0.45 & 0.63 & 0.44 & 0.65 \\

Gemma 3 & 0.0 & 0.0 & 0.2 & 0.1 & 0.2 & 
 0.29 & 0.34 & 0.27 & 0.32 & 0.34 & 
 1.0 & 1.0 & 0.8 & 1.0 & 0.8 & 
 1.0 & 1.0 & 0.78 & 1.0 & 0.76 \\

GPT-4o-mini & 0.0 & 0.2 & 0.0 & 0.0 & 0.0 & 
0.23 & 0.23 & 0.24 & 0.26 & 0.20 & 
1.0 & 1.0 & 0.8 & 0.8 & 0.9 &
0.98 & 0.98 & 0.76 & 0.80 & 0.90 \\

GPT-4o & 0.0 & 0.1 & 0.0 & 0.1 & 0.1 & 
0.27 & 0.16 & 0.20 & 0.21 & 0.23 &
0.9 & 0.8 & 0.8 & 0.9 & 0.9 & 
0.90 & 0.80 & 0.80 & 0.89 & 0.89 \\

Llama 3.3 & 0.0 & 0.3 & 0.3 & 0.1 & 0.3 & 
0.13 & 0.18 & 0.25 & 0.21 & 0.23 & 
0.7 & 0.8 & 0.9 & 0.9 & 0.9 & 
0.60 & 0.66 & 0.78 & 0.82 & 0.77 \\

Mistral & 0.4 & 0.0 & 0.1 & 0.0 & 0.3 & 
0.13 & 0.14 & 0.14 & 0.10 & 0.11 &
0.5 & 0.8 & 0.8 & 0.5 & 0.5 & 
0.36 & 0.61 & 0.53 & 0.39 & 0.42 \\

o3-mini & 0.0 & 0.8 & 0.7 & 0.8 & 0.7 & 
0.24 & 0.21 & 0.21 & 0.16 & 0.12 & 
0.9 & 0.8 & 0.9 & 0.9 & 0.8 & 
0.90 & 0.77 & 0.90 & 0.9 & 0.8 \\

Phi 4 & 0.0 & 0.3 & 0.2 & 0.2 & 0.4 & 
0.06 & 0.12 & 0.08 & 0.20 & 0.18 & 
0.9 & 0.8 & 0.8 & 0.7 & 0.9 & 
0.63 & 0.57 & 0.42 & 0.32 & 0.51 \\

Qwen 3 & 0.1 & 0.4 & 0.5 & 0.3 & 0.3 & 
0.19 & 0.27 & 0.23 & 0.36 & 0.29 & 
0.9 & 1.0 & 0.9 & 0.9 & 0.9 & 
0.9 & 1.0 & 0.87 & 0.9 & 0.9 \\

Qwen 2.5 & 0.0 & 0.1 & 0.0 & 0.1 & 0.0 & 
0.17 & 0.16 & 0.21 & 0.22 & 0.23 & 
1.0 & 1.0 & 1.0 & 0.9 & 0.9 & 
0.78 & 0.85 & 0.88 & 0.81 & 0.87 \\

\hline
\end{tabular}
}
\caption{Performance of different LLM generators in terms of the soft constraint satisfaction, evaluated by Open\framework. While DeepSeek-R1 and GPT-4o-mini performed great in terms of hard constraints, their soft constraint satisfaction is poor, indicating their inability to take user feedback into account.}
\label{tab:open_eval_soft}
\end{table*}

\subsection{Full Results} \label{app:full_results}

As shown in the main results, we opt for the GPT-4o and o3-mini models for LLM supported evaluation dimensions due to their superiority over other models. In our preliminary pipeline experiments, we notice that the faithfulness ratio is the most expensive part of the evaluation. For a single check, it takes 3 abstract + introduction pairs (2 as few-shot examples and 1 for the evaluated citation sentence). In addition, we repeat the checks three times to make a robust evaluation. The total number of evaluations for a single related work section is directly proportional to the number of citations in the related work document. If a sentence includes multiple citations, we implement our evaluation with each respective citation's abstract and introduction. Finally, completion of a single paper takes 5 iterations. This results in a significant cost to evaluate a single generator for a single paper. The overall cost multipliers are as follows:

{\small
\begin{equation*}
N_{ev\_rep} \times |C| \times N_{iter} \times |D| \times N_{runs} \times N_{generator} \times N_{exp\_type}
\end{equation*}}

where $N_{ev\_rep}$, $|C|$, $N_{iter}$, $|D|$, $N_{runs}$, $N_{generator}$, $N_{exp\_type}$ stand for the number of repeated evaluations, the cardinality of the citation set, the number of iterations, the cardinality of the dataset, the number of pipeline runs for a generator, the number of generator models, and the number of experiment types (e.g., full pipeline, introduction of new paper, style changes), respectively. This setup can easily climb up to 5 digit costs. Therefore, we implemented our multiple runs on a smaller subset of papers (10 instances) to diminish estimated cost. We provide the mean and standard deviation of different runs in Tables \ref{tab:gpt-full} - \ref{tab:gemma-style}. 

The hard constraint pass ratio for hallucinated and missing papers is quite high in general but not consistently perfect. For length and citation emphasis, almost all LLMs except o3-mini perform poorly. This fluctuation leads to higher standard deviations. On the other hand, in the LLM based evaluations, we observe lower standard deviation values. The generator behavior is also consistent across different experiment types (e.g., first two iterations before the simulated user inference). After having certainty in different run results, we present full dataset results in main text.

Figure~\ref{fig:across_paper} shows per-paper score distributions for four generators at the final iteration (10-instance subset), complementing the run-to-run standard deviations in Tables~\ref{tab:gpt-full}--\ref{tab:gemma-style}. We show four representative criteria (missing papers, length, citation emphasis, and positioning ratio); saturated criteria (e.g., positioning existence) show negligible spread and are omitted. The spread is substantial: citation emphasis varies widely for every model, Gemma 3 shows the largest variance on missing papers, and length separates models most clearly (o3-mini highest). Positioning ratio is near-saturated for most models, whereas Llama 3.3 varies widely across papers, consistent with its lower average in Table~\ref{tab:llama-full}. This confirms that the reported averages summarize heterogeneous per-paper behavior.

\subsection{Performance Changes Over Iterations}\label{app:exp_delta}

\begin{figure*}
\centering
\includegraphics[width=\linewidth]{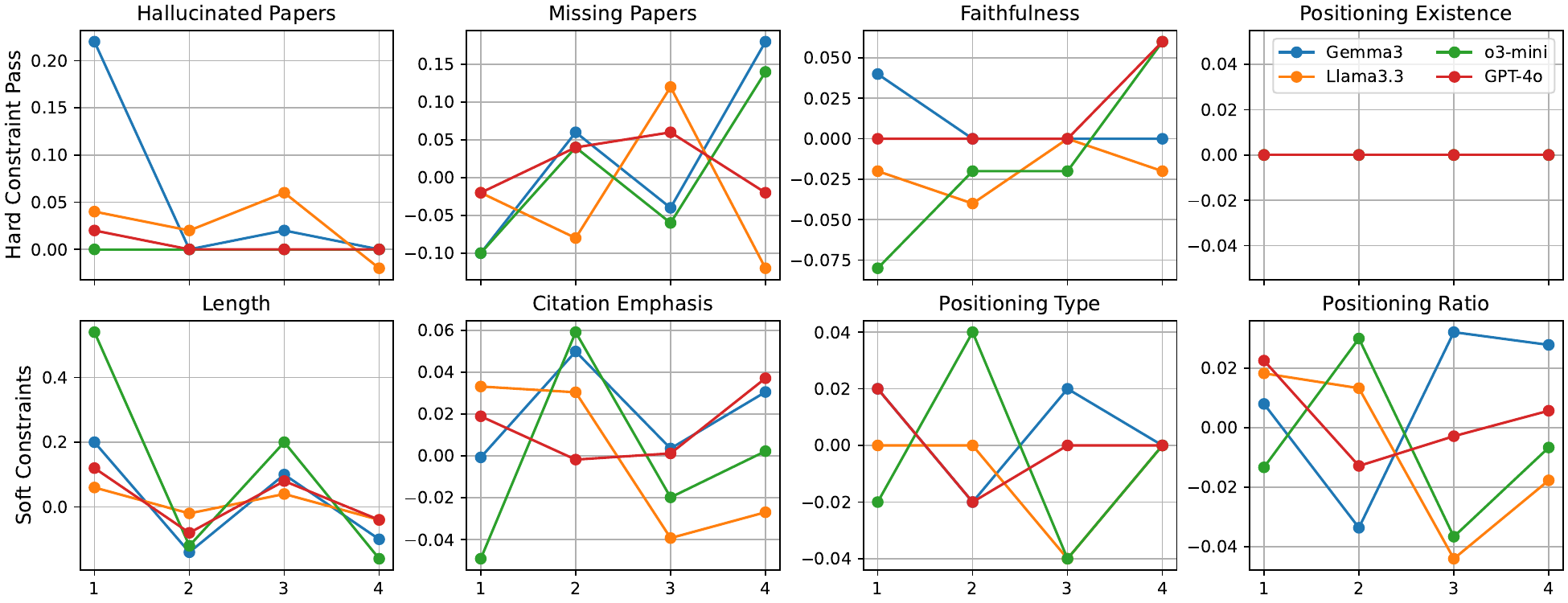}
\caption{Improvements per iteration in hard constraint passing rates and soft constraint passing, evaluated by Precise\framework.}
\label{fig:full_pipeline_delta}
\end{figure*}

\begin{figure*}
\centering
\includegraphics[width=\linewidth]{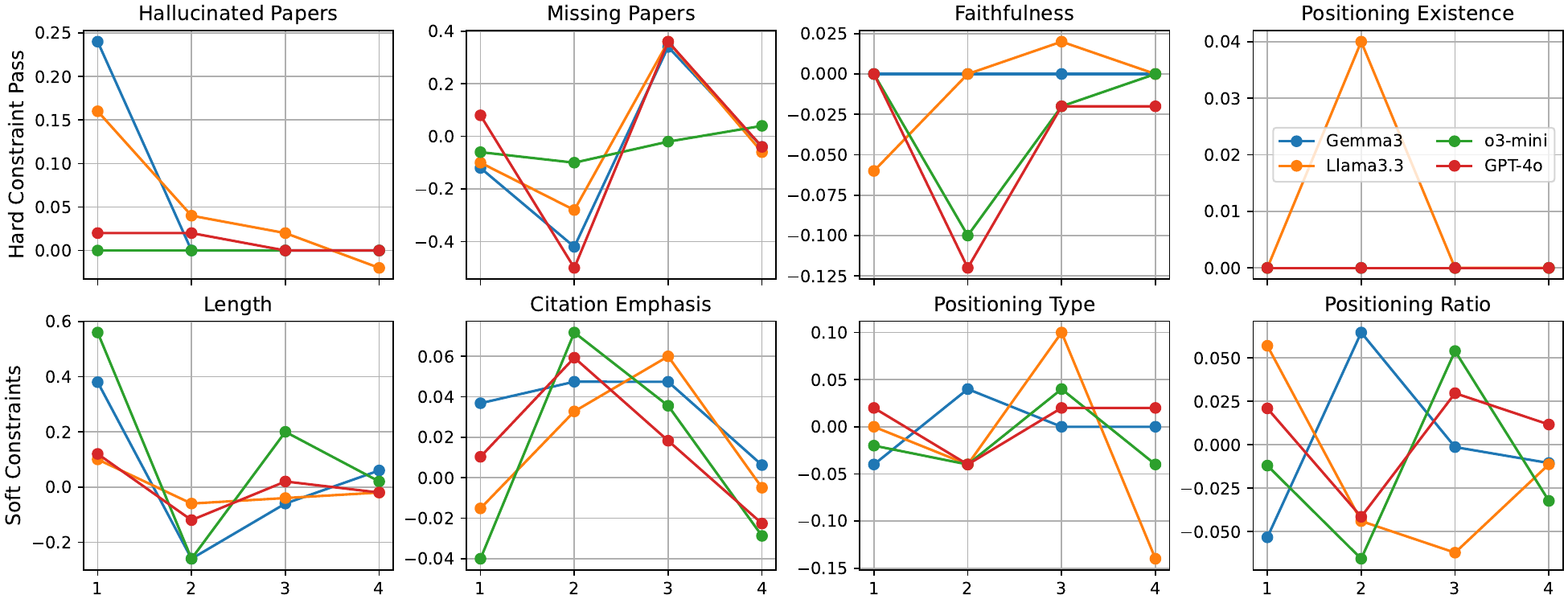}
\caption{Improvements per iteration, with new papers added on 3rd iteration, in hard constraint passing rates and soft constraint passing, evaluated by Precise\framework.}
\label{fig:new_paper_delta}
\end{figure*}

\begin{figure*}
\centering
\includegraphics[width=\linewidth]{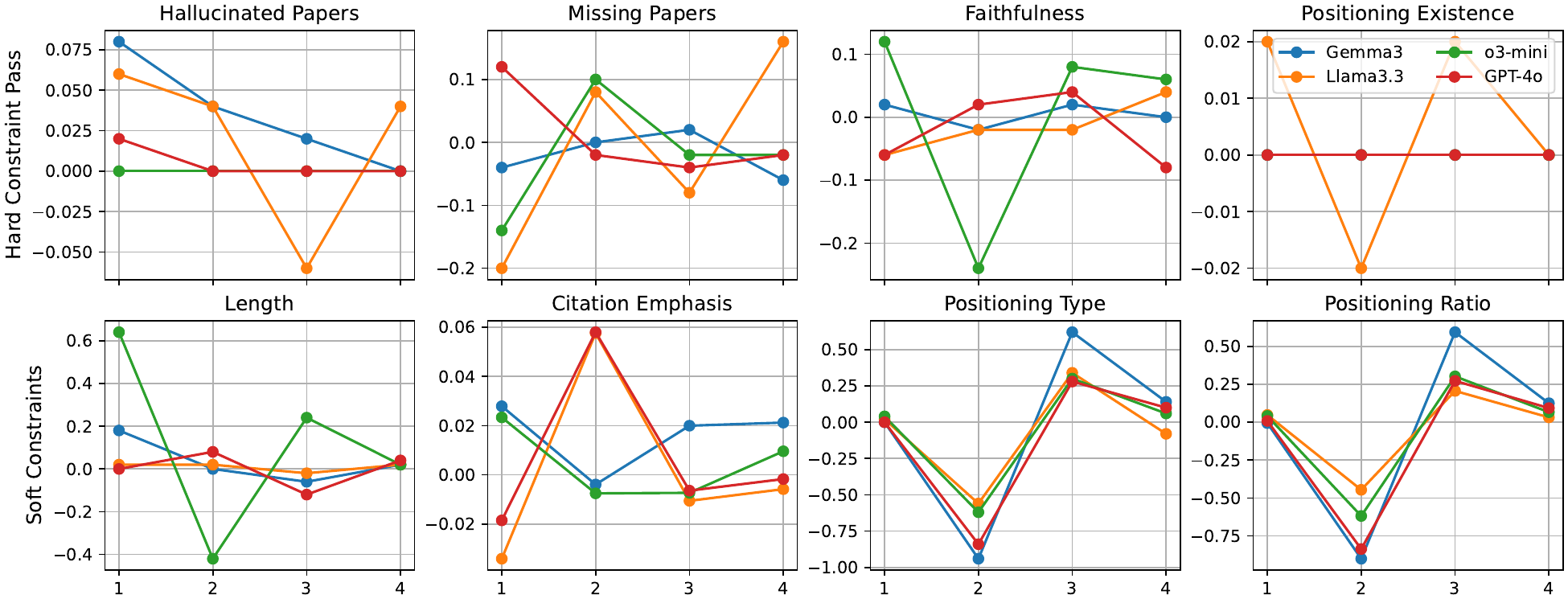}
\caption{Improvements per iteration, with style change introduced on 3rd iteration, in hard constraint passing rates and soft constraint passing, evaluated by Precise\framework.}
\label{fig:style_change_delta}
\end{figure*}

In Figures \ref{fig:full_pipeline_delta}, \ref{fig:new_paper_delta}, and \ref{fig:style_change_delta}, we demonstrate the performance changes across iterations. 

\section{Expert Evaluation}\label{app:human_eval}

Human experts interact with models for three iterations, resulting in a total of 50 expert judgments (15 experts $\times$ 3 iterations $+$ one expert x 5 iterations) for each criterion: citation faithfulness, positioning, and instruction following.

The pairwise comparative strategy is adopted to minimize cognitive burden on participants and subjective direct scoring \cite{Phelps:2015}. Since the number of comparisons increases quadratically ($O(n^2)$) with the number of models, it is not possible to make a complete set of comparisons. Instead, we use the TrueSkill\texttrademark\ algorithm \cite{Herbrich:2006} to dynamically rank the generator models based on expert selections and find the most informative comparison pairs.

For Precise\framework\ and Open\framework, we evaluate both drafts at each iteration and select the higher-scoring one as the better model. For baselines, we selected either higher scoring ones or selection among pairwise comparison depending on the structure of the baseline models. We utilize the improvement between consecutive iterations as a measure of feedback-following.

\begin{figure*}[h]
\centering
\includegraphics[width=1\linewidth]{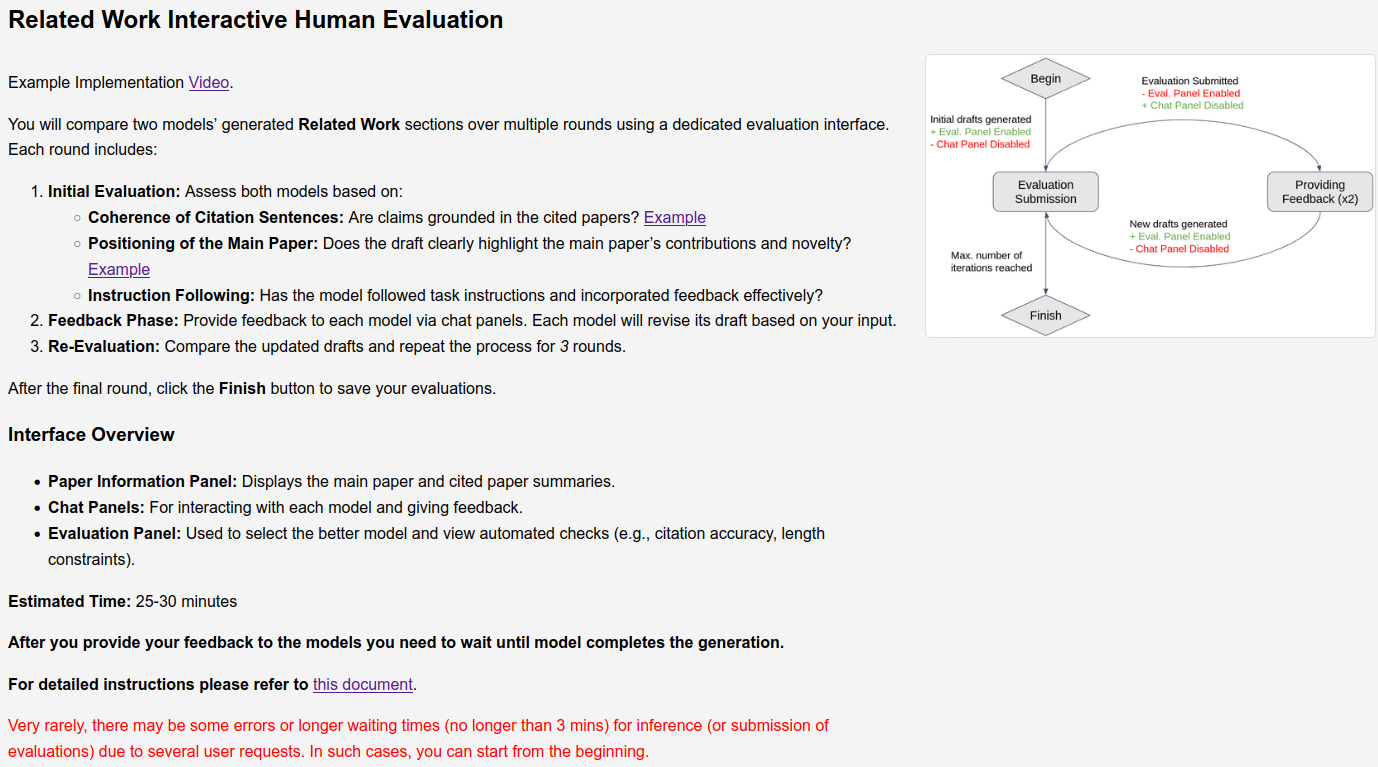}
\caption{Expert Evaluation: Instruction to experts}
\label{fig:humam_eval_ins}
\end{figure*}

\begin{figure*}
\centering
\includegraphics[width=1\linewidth]{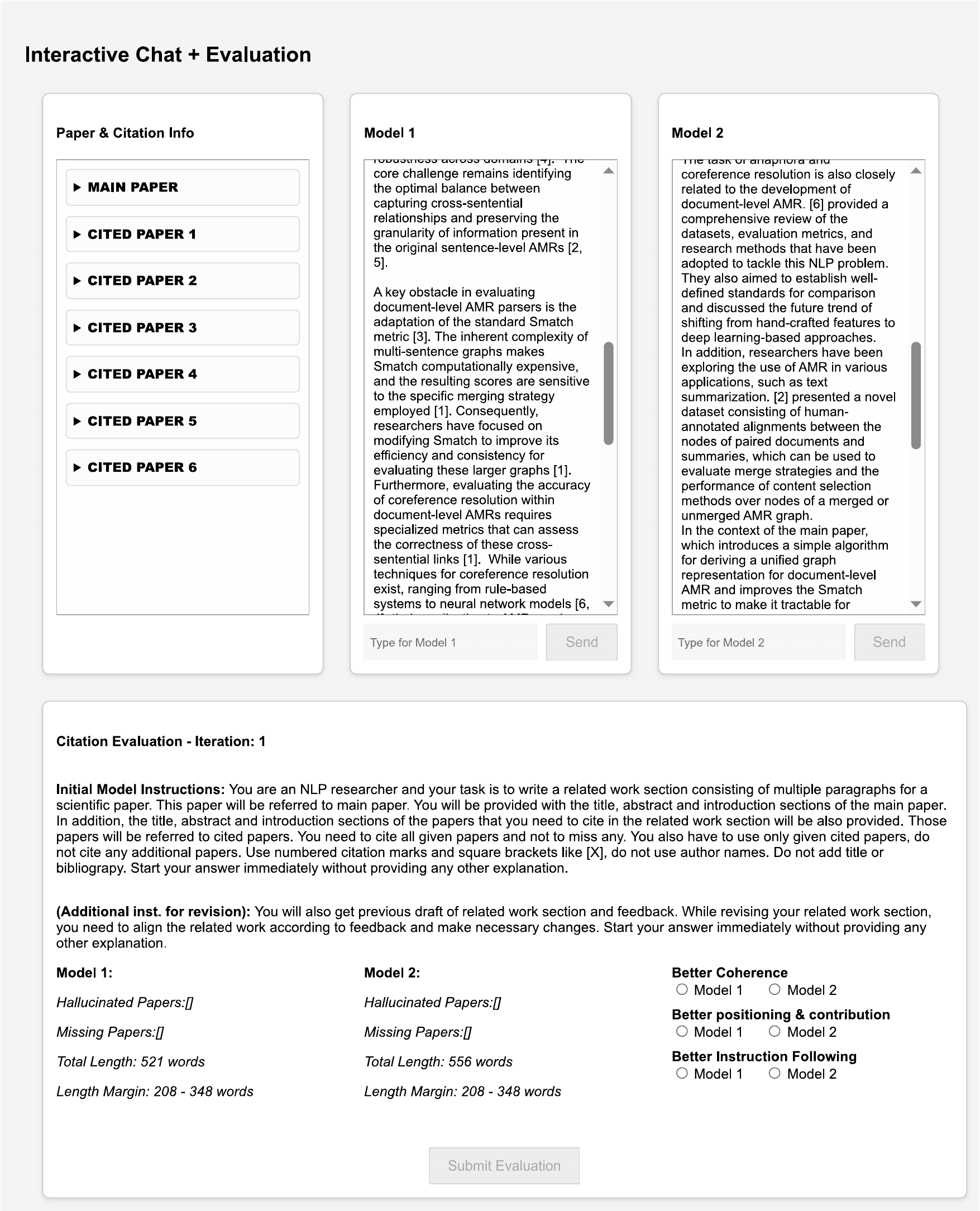}
\caption{Expert Evaluation: Interactive chat and evaluation panels}
\label{fig:humam_eval}
\end{figure*}

Before starting the expert evaluation, we provide participants with a detailed instruction document outlining the user study to participants. This document includes introduction of chat and evaluation panels, explanation of evaluation aspects (e.g., faithfulness, positioning, and feedback following). Screenshot of instructions are provided in Figure \ref{fig:humam_eval_ins}. Each evaluation aspect is complemented with an example to clarify the points that experts should focus on. To reduce the cognitive load, we provide missing and hallucinated paper information along with length evaluation for each generated draft. In addition, we include an instructional video that demonstrates how to interact with the evaluation interface. We present an example visual from the evaluation page in Figure \ref{fig:humam_eval}. 

\begin{table*}
\resizebox{\textwidth}{!}{
\begin{tabular}{l|ccc|ccc}
\hline
\multirow{2}{*}{\textbf{Framework}} & \multicolumn{3}{c|}{\textbf{NLP Domain}} & \multicolumn{3}{c}{\textbf{CV Domain}} \\
\textbf{} & \textbf{F} & \textbf{P} & \textbf{Fb} & \textbf{F} & \textbf{P} & \textbf{Fb} \\
\hline
STE & 0.53 {\scriptsize[0.36, 0.69]} & 0.63 {\scriptsize[0.45, 0.77]} & 0.47 {\scriptsize[0.31, 0.64]} & 0.44 {\scriptsize[0.25, 0.66]} & 0.39 {\scriptsize[0.20, 0.61]} & 0.39 {\scriptsize[0.20, 0.61]} \\
DS-GRM & \underline{0.66} {\scriptsize[0.48, 0.80]} & 0.63 {\scriptsize[0.45, 0.77]} & 0.50 {\scriptsize[0.34, 0.66]} & \underline{0.67} {\scriptsize[0.44, 0.84]} & 0.50 {\scriptsize[0.29, 0.71]} & \textbf{0.67} {\scriptsize[0.44, 0.84]} \\
Nemotron & 0.41 {\scriptsize[0.26, 0.58]} & 0.53 {\scriptsize[0.36, 0.69]} & 0.28 {\scriptsize[0.16, 0.45]} & 0.44 {\scriptsize[0.25, 0.66]} & 0.28 {\scriptsize[0.13, 0.51]} & 0.44 {\scriptsize[0.25, 0.66]} \\
\hline
Open\framework & \underline{0.66} {\scriptsize[0.48, 0.80]} & \underline{0.66} {\scriptsize[0.48, 0.80]} & \underline{0.66} {\scriptsize[0.48, 0.80]} & \underline{0.67} {\scriptsize[0.44, 0.84]} & \textbf{0.72} {\scriptsize[0.49, 0.87]} & \textbf{0.67} {\scriptsize[0.44, 0.84]} \\
Precise\framework & \textbf{0.78} {\scriptsize[0.61, 0.89]} & \textbf{0.75} {\scriptsize[0.58, 0.87]} & \textbf{0.69} {\scriptsize[0.51, 0.82]} & \textbf{0.72} {\scriptsize[0.49, 0.87]} & \textbf{0.72} {\scriptsize[0.49, 0.87]} & \textbf{0.67} {\scriptsize[0.44, 0.84]} \\
\hline
\end{tabular}
}
\caption{Match rate with expert judgments, with Wilson 95\% confidence intervals (NLP $n=32$, CV $n=18$). F: Faithfulness, P: Positioning, Fb: Feedback.}
\label{tab:confidence}
\end{table*}

To measure the alignment with expert selections, we use the scores of the corresponding evaluation modules (i.e., faithfulness, positioning) for the drafts produced by each model at each iteration and select the highest scoring one. Since we do not have an evaluation module that directly overlaps with the general feedback following, we approach the problem from a relative improvement perspective. Since expert instruction or feedback to the model is meant to improve the current status of the draft, we compute improvement by measuring score differences between consecutive iterations. For each evaluation module, we determine which model shows greater improvement. The model that achieves more improvements across modules is considered to have followed feedback more effectively. On the other hand, the Self-taught Evaluator is trained to implement pairwise evaluation. We provide the model with generator drafts for each iteration along with explanations corresponding evaluation perspective. Since the model directly select one of the drafts, no additional processing is needed.

\paragraph{Confidence intervals for expert evaluation.} We report Wilson 95\% confidence intervals for the expert match rates in Table \ref{tab:confidence}. At this sample size ($n=32$ for NLP, $n=18$ for CV), the intervals are wide and partially overlap, so we do not claim statistical significance at the level of individual criteria. Nonetheless, the results show a consistent pattern: Precise\framework\ attains the highest point estimate on every criterion in both domains, and its lower bounds lie at or above most baselines' point estimates. The overall ordering holds across all criteria and both domains. This consistency across two independent domains supports the reliability of \framework's alignment with expert judgments. The sample size reflects the nature of the study: our evaluation relies on domain experts performing multi-turn evaluation, rather than crowd annotation. Recruiting participants at this level of expertise is inherently constrained.

\begin{table*}
\centering
\begin{tabular}{p{\dimexpr\linewidth-2\tabcolsep\relax}}
\hline
{\fontsize{9}{9}\selectfont \textbf{Example 1:}} \\
{\fontsize{9}{9}\selectfont  PAPER CONTEXT: Neural Architecture Search methods are effective but often use complex algorithms to come up with the best architecture. We propose an approach with three basic steps that is conceptually much simpler. First we train N random architectures to generate N (architecture, validation accuracy) pairs and use them to train a regression model that predicts accuracy based on the architecture. Next, we use this regression model to predict the validation accuracies of a large number of random architectures. Finally, we train the top-K predicted architectures and deploy the model with the best validation result. While this approach seems simple, it is more than 20× as sample efficient as Regularized Evolution on the NASBench-101 benchmark and can compete on ImageNet with more complex approaches based on weight sharing, such as ProxylessNAS. The original Neural Architecture Search (NAS) methods have resulted in improved accuracy but they came at a high computational cost [27, 20, 19] . Recent advances have reduced this cost significantly [15, 9, 26, 4, 18, 5, 2, 17, 24, 23, 3] , but many of them require nontrivial specialized implementations. For example, weight sharing introduces additional complexity into the search process, and must be carefully tuned to get good results. With an infinite compute budget, a naive approach to architecture search would be to sample tens or hundreds of thousands of random architectures, train and evaluate each one, and then select the architectures with the best validation set accuracies for deployment; this is a straightforward application of the ubiquitous random search heuristic. However, the computational requirements of this approach makes it infeasible in practice. For example, to exhaustively train and evaluate each of the 400,000 architectures in the NASBench [25] search space, it would take roughly 25 years of TPU training time. Only a small number of companies and corporate research labs can afford this much compute, and it is far out of reach for most ML practitioners. One way to alleviate this is to identify a small subset of promising models. If we can do this with a reasonably high recall (most models selected are indeed of high quality) then we can train and validate just this limited set of models to reliably select a good one for deployment. To achieve this, the proposed Neural Predictor uses the following steps to perform an architecture search: (1) Build a predictor by training N random architectures to obtain N (architecture, validation accuracy) pairs. Use this data to train a regression model. (2) Quality prediction using the regression model over a large set of random architectures. Select the K most promising architectures for final validation. (3) Final validation of the top K architectures by training them. Then we select the model with the highest validation accuracy to deploy. The workflow is illustrated in Figure 1 . In this setup, the first step is a traditional regression problem where we first generate a dataset of N samples to train on. The second step can be carried out efficiently because evaluating a model using the predictor is cheap. The third step is nothing more than traditional validation where we only evaluate a well curated set of K models. While the method outlined above might seem straightforward, it is very effective: The Neural Predictor strongly outperforms random search on NASBench-101. It is also about 22.83 times more sample-efficient than Regularized Evolution, the best performing method in the NASBench-101 paper. The Neural Predictor can easily handle different search spaces. In addition to NASBench-101, we evaluated it on the ProxylessNAS [4] search space and found that the predicted architecture is as accurate as Proxyless-NAS and clearly better than random search. The architecture selection process uses two of the most ubiquitous tools from the ML toolbox: random sampling and supervised learning. In contrast, many existing NAS approaches rely on reinforcement learning, weight sharing, or Bayesian optimization. The most computationally intensive components of the proposed method (training N models in step 1 and K models in step 3) are highly parallelizable when sufficient computation resources are available.} \\
{\fontsize{9}{9}\selectfont CITATION SENTENCE: Additionally, simple predictors that regress on features extracted from architectures, such as textual encoding schemes or direct structural descriptions, have been found effective for performance approximation, enabling significant sample efficiency improvements [9][10].} \\ \\
{\fontsize{9}{9}\selectfont CITED PAPER: 10} \\ \\
{\fontsize{9}{9}\selectfont REASONING: The context discusses the effectiveness of a Neural Predictor method that uses regression modeling to predict validation accuracy based on architecture features, achieving significant sample efficiency improvements over traditional approaches. The citation sentence aligns with the context in pointing out the effectiveness of simple predictors for performance approximation, which correlates with the regression steps described in the Neural Predictor method. Therefore, the citation sentence is entailed by the context and follows from it. Therefore, the answer should be "yes".} \\ \\
{\fontsize{9}{9}\selectfont ANSWER: Yes} \\
\hline
\end{tabular}
\caption{Contrastive few-shot first example (positive) for faithfulness evaluation}
 \label{tab:coherence_example_1}
\end{table*}

\begin{table*}
\centering
\begin{tabular}{p{\dimexpr\linewidth-2\tabcolsep\relax}}
\hline
{\fontsize{9}{9}\selectfont \textbf{Example 2:}} \\
{\fontsize{9}{9}\selectfont  PAPER CONTEXT: We present a dialogue generation model that directly captures the variability in possible responses to a given input, which reduces the boring output issue of deterministic dialogue models. Experiments show that our model generates more diverse outputs than baseline models, and also generates more consistently acceptable output than sampling from a deterministic encoder-decoder model. The task of open-domain dialogue generation is an area of active development, with neural sequence-to-sequence models dominating the recently published literature (Shang et al., 2015; Vinyals and Le, 2015; Li et al., 2016b,a; Serban et al., 2016). Most previously published models train to minimise the negative log-likelihood of the training data, and then at generation time either perform beam search to find the output Y which maximises P (Y |input) (Shang et al., 2015; Vinyals and Le, 2015; Serban et al., 2016 ) (ML decoding), or sample from the resulting distribution (Serban et al., 2016). A notorious issue with ML decoding is that this tends to generate short, boring responses to a wide range of inputs, such as "I don't know". These responses are common in the training data, and can be replies to a wide range of inputs (Li et al., 2016a; Serban et al., 2016) . In addition, shorter responses typically have higher likelihoods, and so wide beam sizes often result in very short responses (Tu et al., 2017; Belz, 2007) . To resolve this problem, Li et al. (2016a) propose instead using maximum mutual information with a length boost as a decoding objective, and report more interesting generated responses. Further, natural dialogue is not deterministic; for example, the replies to "What's your name and where do you come from?" will vary from person to person. Li et al. (2016b) have proposed learning representations of personas to account for interperson variation, but there can be variation even among a single person's responses to certain questions. Recently, Serban et al. (2017) have introduced latent variables to the dialogue modelling framework, to model the underlying distribution over possible responses directly. These models have the benefit that, at generation time, we can sample a response from the distribution by first sampling an assignment of the latent variables, and then decoding deterministically. In this way, we introduce stochasticity without resorting to sampling from the decoder, which can lead to incoherent output -see Table 1 for examples. In this paper, we present a latent variable model for one-shot dialogue response, and investigate what kinds of diversity the latent variables capture. Our experiments show that our model has higher lexical as well as sentential diversity than baseline models. We also show that our model generates more acceptable diverse output than sampling from a deterministic decoder. We end by noting that all three methods proposed above to combat the 'maximum likelihood response' are ways of decreasing the probability of the generated output, and report some preliminary results for how response probability interacts with grammaticality and interestingness.} \\ \\
{\fontsize{9}{9}\selectfont CITATION SENTENCE: Natural Language Inference has emerged as an essential tool for modeling semantic relationships, with applications extending from textual entailment assessments [11,12] to dialogue consistency verification [17].} \\ \\
{\fontsize{9}{9}\selectfont CITED PAPER: 11} \\ \\
{\fontsize{9}{9}\selectfont REASONING: The citation sentence discusses Natural Language Inference (NLI) and its applications, which is not directly related or entailed by the presented context on dialogue generation models and their improvements. The context does not specifically mention or address NLI or its use in dialogue consistency verification. As such, the hypothesis cannot be inferred or entailed from the given context. Therefore, the answer should be "no".} \\ \\
{\fontsize{9}{9}\selectfont ANSWER: No} \\
\hline
\end{tabular}
\caption{Contrastive few-shot second example (negative) for faithfulness evaluation}
 \label{tab:coherence_example_2}
\end{table*}

\begin{table*}
\centering
\begin{tabular}{p{\dimexpr\linewidth-2\tabcolsep\relax}}
\hline
{\fontsize{9}{9}\selectfont \textbf{Example 1:}} \\
{\fontsize{9}{9}\selectfont  DRAFT: Previous studies on TDMR (Task-Dataset-Metric-Result) extraction primarily rely on data sources such as NLP-progress or PapersWithCode. While these sources are popular, they lack rigorous quality control, including standardized TDM entity representation and complete publication coverage. Some efforts, such as [1] and [2], go further by extracting TDM triples and result values, and normalizing them for leaderboard construction. However, these approaches are constrained by their reliance on a closed domain, requiring extracted triples to match a pre-defined TDM taxonomy. Other studies, such as [3] and [4], only extract TDM triples without result values, thereby forgoing leaderboard construction. Meanwhile, [6] focuses on extracting result values, but only in the context of pre-defined TDM triples, making it similar in limitation to [1]. Consequently, these methods are not adaptable to the dynamic nature of emerging benchmarks, where new tasks, datasets, and metrics are frequently introduced.
In a parallel research thread, scientific leaderboards have been treated as a form of scientific knowledge graphs composed of Task, Dataset, and Metric entities. Information extraction in this context has been widely studied, as seen in works such as [5, 7, 8, 9, 10]. These studies typically focus on linking scientific content to structured entities but often assume that the schema or triples are known in advance.
In contrast to prior work, our approach addresses several critical limitations. We construct our dataset manually from publications to ensure comprehensive TDMR annotations, rather than relying on incomplete or unstandardized third-party sources. We normalize TDMR tuples for accurate leaderboard construction and propose novel experimental settings that reflect realistic, dynamic scenarios, including cold-start conditions without any pre-defined TDM taxonomy. By simultaneously constructing a scientific knowledge graph and linking publications to it in an open-domain setting, our method is inherently adaptable to the evolving research landscape.} \\ \\
{\fontsize{9}{9}\selectfont REASONING: The related work draft does not mention the main paper's contributions and its position in each individual paragraph; instead, it waits until the final paragraph to outline the specific contributions of the study. The contributions are clearly summarized and detailed only in the final paragraph. Therefore, the answer should be "2".} \\ \\
{\fontsize{9}{9}\selectfont ANSWER: 2} \\
\hline
\end{tabular}
\caption{Contrastive few-shot first example (final paragraph contribution) for positioning type evaluation}
 \label{tab:cont_type_example_1}
\end{table*}

\begin{table*}
\centering
\begin{tabular}{p{\dimexpr\linewidth-2\tabcolsep\relax}}
\hline
{\fontsize{9}{9}\selectfont \textbf{Example 2:}} \\
{\fontsize{9}{9}\selectfont DRAFT: In terms of data source, previous studies use either NLP-progress or paperswithcode. These sources, however, lack rigorous quality assurance, such as standardizing TDM entities across different leaderboards and ensuring complete coverage of relevant publications. Similar to our work, [1] and [2] extract TDM triples along with the results values and apply normalization for leaderboard construction. However, both studies assume a closed domain and match extracted TDM triples to a pre-defined TDM triple taxonomy. On the other hand, some studies only partially extract TDMR tuples and do not apply normalization. For example, [3] and [4] extract TDM triples without results. Therefore, these works do not deal with leaderboard construction. In addition, [6] extract the results values depending on the pre-defined TDM triples. Both [3] and [6] leverage pre-defined TDM triples in an extraction process similar to [1]. Since these approaches require a pre-defined taxonomy of TDM triples, they are incompatible with a realistic task definition. In short, none of the previous work is adaptable to the constantly emerging benchmarks driven by new research and innovation. In this work, we address the aforementioned problems. Unlike previous work, we (1) manually construct our dataset directly from publications to ensure complete TDMR annotations, (2) apply normalization for leaderboard construction, and (3) propose different experimental settings to simulate real-world scenarios. Part of the scientific leaderboards can be viewed as a special type of scientific knowledge graph that includes three types of entities (Task, Dataset, Metric) and the relations between them, which have been the primary focus of the previous studies on information extraction from scientific literature [5, 7, 8, 9, 10]. Our work in the cold start scenario, in which we do not assume any pre-defined TDM triple is given, constructs such a scientific knowledge graph and links the papers to the nodes in the graph simultaneously.} \\ \\
{\fontsize{9}{9}\selectfont REASONING: Contributions and position of the paper in the literature are mentioned within each paragraph rather than being summarized at the end. The first paragraph discusses previous work and their limitations, then explicitly states what the current work does differently, listing three specific contributions. The second paragraph again discusses previous studies but also explains how the current work differs by constructing a knowledge graph without assuming predefined triples and linking papers to nodes. Therefore, the answer should be "1".} \\ \\
{\fontsize{9}{9}\selectfont ANSWER: 1} \\
\hline
\end{tabular}
\caption{Contrastive few-shot second example (each paragraph contribution) for positioning type evaluation}
 \label{tab:cont_type_example_2}
\end{table*}

\begin{table*}
\centering
\begin{tabular}{p{\dimexpr\linewidth-2\tabcolsep\relax}}
\hline
{\fontsize{9}{9}\selectfont \textbf{Example 3:}} \\
{\fontsize{9}{9}\selectfont DRAFT: Several efforts have been made to automate the extraction and organization of performance results from scientific publications. Early work by [1] introduced a framework, TDMS-IE, for identifying task, dataset, metric, and score tuples from NLP papers, establishing a foundation for automatic leaderboard construction. Similarly, AxCell [2] proposes a robust pipeline that utilizes structural analysis and novel table segmentation techniques to extract results from machine learning papers, demonstrating significant improvements over prior methods. ORKG-Leaderboards [3] offers a systematic approach that integrates leaderboard extraction into a knowledge graph framework, enabling machine-actionable publishing and dynamic visualization of state-of-the-art performance. TELIN [4] focuses on extracting leaderboard-relevant entities from PDFs using a semi-automated approach that reduces human annotation needs through targeted entity refinement.
In addition to these extraction systems, several datasets and benchmarks have been introduced to facilitate the development and evaluation of leaderboard construction tools. LEGOBench [5] provides a large-scale benchmark derived from arXiv and PapersWithCode, and evaluates both language model-based and graph-based approaches. SciERC and its associated framework SciIE [6] support multi-task extraction of entities and relations, enabling construction of scientific knowledge graphs. SciREX [7] extends information extraction to the document level, capturing relationships that span across sections, which is critical for leaderboard generation from full papers.
Other contributions include TDMSci [8], a specialized corpus annotated with task, dataset, and metric entities, which supports the development of more accurate extraction models. SciNLP-KG [9] introduces methods for extracting entity relations from NLP literature to build a knowledge graph, highlighting its potential use in leaderboard automation. Additionally, a diachronic analysis of NLP research trends by [10] shows the evolving influence of tasks, methods, and datasets, providing a broader context for understanding the dynamic nature of scientific benchmarks.} \\ \\
{\fontsize{9}{9}\selectfont REASONING: The related work section draft provided does not explicitly mention or discuss the main paper's contributions or its position. Each paragraph focuses on summarizing existing research efforts and methodologies without indicating how the current paper builds upon or differs from these works. Additionally, the final paragraph does not serve as a summary of the main paper's contributions; instead, it continues to discuss other related works without tying them back to the current study's advancements. Therefore, the answer should be "3".} \\ \\
{\fontsize{9}{9}\selectfont ANSWER: 3} \\
\hline
\end{tabular}
\caption{Contrastive few-shot third example (each paragraph contribution) for positioning type evaluation}
 \label{tab:cont_type_example_3}
\end{table*}

\begin{table*}
\centering
{\fontsize{9}{9}\selectfont
\begin{tabular}{p{\dimexpr\linewidth-2\tabcolsep\relax}}
\hline
\textbf{Example 1:} \\
DRAFT: In terms of data source, previous studies use either NLP-progress or paperswithcode. These sources, however, lack rigorous quality assurance, such as standardizing TDM entities across different leaderboards and ensuring complete coverage of relevant publications. Similar to our work, [1] and [2] extract TDM triples along with the results values and apply normalization for leaderboard construction. However, both studies assume a closed domain and match extracted TDM triples to a pre-defined TDM triple taxonomy. On the other hand, some studies only partially extract TDMR tuples and do not apply normalization. For example, [3] and [4] extract TDM triples without results. Therefore, these works do not deal with leaderboard construction. In addition, [6] extract the results values depending on the pre-defined TDM triples. Both [3] and [6] leverage pre-defined TDM triples in an extraction process similar to [1]. Since these approaches require a pre-defined taxonomy of TDM triples, they are incompatible with a realistic task definition. In short, none of the previous work is adaptable to the constantly emerging benchmarks driven by new research and innovation. In this work, we address the aforementioned problems. Unlike previous work, we (1) manually construct our dataset directly from publications to ensure complete TDMR annotations, (2) apply normalization for leaderboard construction, and (3) propose different experimental settings to simulate real-world scenarios. Part of the scientific leaderboards can be viewed as a special type of scientific knowledge graph that includes three types of entities (Task, Dataset, Metric) and the relations between them, which have been the primary focus of the previous studies on information extraction from scientific literature [5, 7, 8, 9, 10]. Our work in the cold start scenario, in which we do not assume any pre-defined TDM triple is given, constructs such a scientific knowledge graph and links the papers to the nodes in the graph simultaneously. \\[3pt]
REASONING: The draft states the main paper's contribution and how it differs from existing literature. It outlines the limitations of previous studies and then explicitly states how the current work addresses these issues through specific contributions, such as dataset construction and handling cold start scenarios without pre-defined TDM triples. Therefore, the answer should be ``yes''. \\[3pt]
ANSWER: Yes \\
\hline
\textbf{Example 2:} \\
DRAFT: Several efforts have been made to automate the extraction and organization of performance results from scientific publications. Early work by [1] introduced a framework, TDMS-IE, for identifying task, dataset, metric, and score tuples from NLP papers, establishing a foundation for automatic leaderboard construction. Similarly, AxCell [2] proposes a robust pipeline that utilizes structural analysis and novel table segmentation techniques to extract results from machine learning papers, demonstrating significant improvements over prior methods. ORKG-Leaderboards [3] offers a systematic approach that integrates leaderboard extraction into a knowledge graph framework, enabling machine-actionable publishing and dynamic visualization of state-of-the-art performance. TELIN [4] focuses on extracting leaderboard-relevant entities from PDFs using a semi-automated approach that reduces human annotation needs through targeted entity refinement. In addition to these extraction systems, several datasets and benchmarks have been introduced to facilitate the development and evaluation of leaderboard construction tools. LEGOBench [5] provides a large-scale benchmark derived from arXiv and PapersWithCode, and evaluates both language model-based and graph-based approaches. SciERC and its associated framework SciIE [6] support multi-task extraction of entities and relations, enabling construction of scientific knowledge graphs. SciREX [7] extends information extraction to the document level, capturing relationships that span across sections, which is critical for leaderboard generation from full papers. Other contributions include TDMSci [8], a specialized corpus annotated with task, dataset, and metric entities, which supports the development of more accurate extraction models. SciNLP-KG [9] introduces methods for extracting entity relations from NLP literature to build a knowledge graph, highlighting its potential use in leaderboard automation. Additionally, a diachronic analysis of NLP research trends by [10] shows the evolving influence of tasks, methods, and datasets, providing a broader context for understanding the dynamic nature of scientific benchmarks. \\[3pt]
REASONING: The draft provides a comprehensive overview of related work but does not explicitly or implicitly mention the main paper's own contribution or how it differs from the existing literature. It talks about the efforts made but doesn't bridge into how this work builds upon them or addresses gaps they left. It focuses solely on summarizing prior work without establishing the unique position or advancement made by the current study. Therefore, the answer should be ``no''. \\[3pt]
ANSWER: No \\
\hline
\end{tabular}%
}
\caption{Contrastive few-shot examples for positioning ratio evaluation of each paragraph.}
\label{tab:cont_direct_eval_examples}
\end{table*}

\begin{table*}
\centering
\begin{tabular}{p{\dimexpr\linewidth-2\tabcolsep\relax}}
\hline
{\fontsize{9}{9}\selectfont \textbf{Example 1:}} \\
{\fontsize{9}{9}\selectfont CONTEXT: In a parallel research thread, scientific leaderboards have been treated as a form of scientific knowledge graphs composed of Task, Dataset, and Metric entities. Information extraction in this context has been widely studied, as seen in works such as [5, 7, 8, 9, 10]. These studies typically focus on linking scientific content to structured entities but often assume that the schema or triples are known in advance.} \\ \\
{\fontsize{9}{9}\selectfont FINAL: In contrast to prior work, our approach addresses several critical limitations. We construct our dataset manually from publications to ensure comprehensive TDMR annotations, rather than relying on incomplete or unstandardized third-party sources. We normalize TDMR tuples for accurate leaderboard construction and propose novel experimental settings that reflect realistic, dynamic scenarios, including cold-start conditions without any pre-defined TDM taxonomy. By simultaneously constructing a scientific knowledge graph and linking publications to it in an open-domain setting, our method is inherently adaptable to the evolving research landscape.} \\ \\
{\fontsize{9}{9}\selectfont REASONING: The context paragraph discusses prior work on scientific leaderboards as knowledge graphs with entities like Task, Dataset, and Metric (TDM). It highlights that prior studies focus on information extraction and linking to structured entities but usually assume a pre-existing schema or set of triples. The final paragraph contrasts the main paper's contributions with this prior work by constructing a scientific knowledge graph and linking publications to it in an open-domain and adaptable manner, which directly relates to the context's theme of schema-free extraction and linking. The final paragraph clearly positions the paper among existing literature and addresses the specific limitation mentioned in the context—i.e., the assumption of known schema—by proposing methods that work without one. Therefore, the answer should be "yes".} \\ \\
{\fontsize{9}{9}\selectfont ANSWER: Yes} \\
{\fontsize{9}{9}\selectfont \textbf{Example 2:}} \\
{\fontsize{9}{9}\selectfont CONTEXT: In a parallel research thread, scientific leaderboards have been treated as a form of scientific knowledge graphs composed of Task, Dataset, and Metric entities. Information extraction in this context has been widely studied, as seen in works such as [5, 7, 8, 9, 10]. These studies typically focus on linking scientific content to structured entities but often assume that the schema or triples are known in advance.} \\ \\
{\fontsize{9}{9}\selectfont FINAL: Building on existing lines of research, this work explores TDMR extraction and scientific leaderboard construction. The dataset is compiled from scientific publications. The approach is designed to accommodate variations in how benchmark results are reported.} \\ \\
{\fontsize{9}{9}\selectfont REASONING: The context paragraph discusses prior research treating scientific leaderboards as knowledge graphs with a focus on information extraction where the schema is often assumed to be known. The main point raised is that existing methods typically presume a known structure for the extracted information. While the final paragraph mentions the extraction of TDMR and the construction of leaderboards—aligning with the topic in the context—it does not explicitly address whether this work assumes a known schema or introduces any method for schema discovery or flexibility. Thus, the final paragraph does not clearly state how it handles the assumption of a known schema, which is a key point of the context paragraph. Therefore, the answer should be "no".} \\ \\
{\fontsize{9}{9}\selectfont ANSWER: No} \\
\hline
\end{tabular}
\caption{Contrastive few-shot examples for positioning ratio evaluation via pairwise comparison between context related work paragraph and final related work paragraph.}
 \label{tab:cont_pairwise_eval_examples}
\end{table*}

\begin{table*}
\centering
\begin{tabular}{p{\dimexpr\linewidth-2\tabcolsep\relax}}
\hline
{\fontsize{9}{9}\selectfont \textbf{SYSTEM:}} \\ 
{\fontsize{9}{9}\selectfont You are an NLP researcher and your task is to write a related work section consisting of multiple paragraphs for a scientific paper. This paper will be referred to main paper. You will be provided with the title, abstract and introduction sections of the main paper. In addition, the title, abstract and introduction sections of the papers that you need to cite in the related work section will be also provided. Those papers will be referred to cited papers. \texttt{\{contribution information\}} You need to cite all given papers and not to miss any. You also have to use only given cited papers, do not cite any additional papers. Use numbered citation marks and square brackets like [X], do not use author names. Do not add title or bibliography. Start your answer immediately without providing any other explanation.} \\ \\
{\fontsize{9}{9}\selectfont \textbf{USER:}} \\
{\fontsize{9}{9}\selectfont MAIN PAPER TITLE: \texttt{\{Title of main paper\}}} \\ \\
{\fontsize{9}{9}\selectfont MAIN PAPER ABSTRACT: \texttt{\{Abstract of main paper\}}} \\ \\
{\fontsize{9}{9}\selectfont MAIN PAPER INTRODUCTION: \texttt{\{Introduction of main paper\}}} \\ \\
{\fontsize{9}{9}\selectfont CITED PAPER [X] TITLE: \texttt{\{Title of cited paper [X]\}}} \\ \\
{\fontsize{9}{9}\selectfont CITED PAPER [X] ABSTRACT: \texttt{\{Abstract of cited paper [X]\}}} \\ \\
{\fontsize{9}{9}\selectfont CITED PAPER [X] INTRODUCTION: \texttt{\{Introduction of cited paper [X]\}}} \\ \\
... \\
\hline
\end{tabular}
\caption{Prompt of draft generation for first iteration}
 \label{tab:first_draft_prompt}
\end{table*}

\begin{table*}
\centering
\begin{tabular}{p{\dimexpr\linewidth-2\tabcolsep\relax}}
\hline
{\fontsize{9}{9}\selectfont \textbf{SYSTEM:}} \\
{\fontsize{9}{9}\selectfont You are an NLP researcher and your task is to revise a related work section consisting of multiple paragraphs for a scientific paper. This paper will be referred to main paper. You will be provided with the title, abstract and introduction sections of the main paper. In addition, the title, abstract and introduction sections of the papers that you need to cite in the related work section will be also provided. Those papers will be referred to as cited papers. \texttt{\{contribution information\}} You need to cite all given papers and not to miss any. You also have to use only given cited papers, do not cite any additional papers. Use numbered citation mark and square brackets like [X], do not use author names. Do not add title or bibliograpy. You will also get previous draft of related work section and feedback. While revising your related work section, you need to align the related work according to feedback and make necessary changes. Start your answer immediately without providing any other explanation.} \\ \\
{\fontsize{9}{9}\selectfont \textbf{USER:}} \\
{\fontsize{9}{9}\selectfont MAIN PAPER TITLE: \texttt{\{Title of main paper\}}} \\ \\
{\fontsize{9}{9}\selectfont MAIN PAPER ABSTRACT: \texttt{\{Abstract of main paper\}}} \\ \\
{\fontsize{9}{9}\selectfont MAIN PAPER INTRODUCTION: \texttt{\{Introduction of main paper\}}} \\ \\
{\fontsize{9}{9}\selectfont CITED PAPER [X] TITLE: \texttt{\{Title of cited paper [X]\}}} \\ \\
{\fontsize{9}{9}\selectfont CITED PAPER [X] ABSTRACT: \texttt{\{Abstract of cited paper [X]\}}} \\ \\
{\fontsize{9}{9}\selectfont CITED PAPER [X] INTRODUCTION: \texttt{\{Introduction of cited paper [X]\}}} \\ \\
... \\ \\
{\fontsize{9}{9}\selectfont PREVIOUS DRAFT: \texttt{\{Generated draft in previous iteration\}}} \\ \\
{\fontsize{9}{9}\selectfont FEEDBACK: \texttt{\{Generated feedback for the previous draft\}}} \\ \\
\hline
\end{tabular}
\caption{Prompt of draft generation after first iteration}
 \label{tab:next_draft_prompt}
\end{table*}

\begin{table*}
\centering
\begin{tabular}{p{\dimexpr\linewidth-2\tabcolsep\relax}}
\hline
{\fontsize{9}{9}\selectfont \textbf{SYSTEM:}} \\
{\fontsize{9}{9}\selectfont You will receive an evaluation report about a related work section draft for a scientific paper. You task is to generate feedback based on this evaluation report. The evaluation report includes (1) missed and hallucinated paper numbers, (2) length of section, (3) evaluation of how much emphasis is placed on each cited paper, (4) sentences lacking faithfulness (5) intended contribution type and evaluation of draft's contribution type. Your feedback should synthesize the items in report into a short concise feedback that explains what should be maintained, improved or revised in the next iteration. It should not be too wordy. Start your answer immediately without providing any other explanation.} \\ \\
{\fontsize{9}{9}\selectfont \textbf{USER:}} \\
{\fontsize{9}{9}\selectfont EVALUATION REPORT: \texttt{\{Evaluation Report\}}} \\ \\
\hline
\end{tabular}
\caption{Prompt of feedback generation}
 \label{tab:feedback_prompt}
\end{table*}

\begin{table*}
\centering
\begin{tabular}{p{\dimexpr\linewidth-2\tabcolsep\relax}}
\hline
{\fontsize{9}{9}\selectfont \textbf{SYSTEM:}} \\ 
{\fontsize{9}{9}\selectfont You will receive some content from a scientific paper, a sentence that is supposed to cite that paper and a specific citation number. Your task is to determine whether the given paper context supports (entails) the sentence for that specific citation number. In cases where more than one paper is referenced in the sentence, as long as context in which given citation number fits the paper content, it should be count as entailment as well. In multiple citation cases, the paper does not have to entail whole sentence. Some examples showing the implementation of the task will be provided. By utilizing the examples, first provide your reasoning, and then your answer. If the paper context entails the citation sentence, answer "yes". If not, answer "no". Your output should be in JSON format.} \\ \\
{\fontsize{9}{9}\selectfont \textbf{USER:}} \\
{\fontsize{9}{9}\selectfont $<$START OF EXAMPLE 1$>$} \\
{\fontsize{9}{9}\selectfont \texttt{\{Example 1\}}} \\
{\fontsize{9}{9}\selectfont $<$END OF EXAMPLE 1$>$} \\ \\  
{\fontsize{9}{9}\selectfont $<$START OF EXAMPLE 2$>$} \\
{\fontsize{9}{9}\selectfont \texttt{\{Example 2\}}} \\
{\fontsize{9}{9}\selectfont $<$END OF EXAMPLE 2$>$} \\ \\  
{\fontsize{9}{9}\selectfont PAPER CONTEXT: \texttt{\{Cited paper abstract and introduction\}}} \\ \\
{\fontsize{9}{9}\selectfont CITATION SENTENCE: \texttt{\{Citation sentence\}}} \\ \\
{\fontsize{9}{9}\selectfont CITATION PAPER: \texttt{\{Number of the cited paper\}}} \\ 
\hline
\end{tabular}
\caption{Faithfulness: System prompts and and contrastive few-shot examples presented in Tables \ref{tab:coherence_example_1}, \ref{tab:coherence_example_2}.}
 \label{tab:coherence_prompt}
\end{table*}

\begin{table*}
\centering
\begin{tabular}{p{\dimexpr\linewidth-2\tabcolsep\relax}}
\hline
{\fontsize{9}{9}\selectfont \textbf{SYSTEM:}} \\
{\fontsize{9}{9}\selectfont You will be given a related work section draft for an academic paper. Your task is to determine whether this draft (1) states the main paper's contribution or its position among the literature in each paragraph, or (2) provides the contributions and/or position in the final paragraph as a summary, or (3) does not mention any contributions/position at all. In addition, some examples showing the implementation of the task will be provided. By utilizing the examples, first provide your reasoning, and then your answer. Your answer should be either "1" for each paragraph, "2" for final paragraph, "3" for no contribution. Your output should be in JSON format.} \\ \\
{\fontsize{9}{9}\selectfont \textbf{USER:}} \\
{\fontsize{9}{9}\selectfont $<$START OF EXAMPLE 1$>$} \\
{\fontsize{9}{9}\selectfont \texttt{\{Example 1\}}} \\
{\fontsize{9}{9}\selectfont $<$END OF EXAMPLE 1$>$} \\ \\  
{\fontsize{9}{9}\selectfont $<$START OF EXAMPLE 2$>$} \\
{\fontsize{9}{9}\selectfont \texttt{\{Example 2\}}} \\
{\fontsize{9}{9}\selectfont $<$END OF EXAMPLE 2$>$} \\ \\
{\fontsize{9}{9}\selectfont $<$START OF EXAMPLE 3$>$} \\
{\fontsize{9}{9}\selectfont \texttt{\{Example 3\}}} \\
{\fontsize{9}{9}\selectfont $<$END OF EXAMPLE 3$>$} \\ \\  
{\fontsize{9}{9}\selectfont DRAFT: \texttt{\{Generated related work section draft\}}} \\
\hline
\end{tabular}
\caption{Positioning existence and positioning type: System prompts and contrastive few-shot examples presented in Tables \ref{tab:cont_type_example_1}, \ref{tab:cont_type_example_2}, and \ref{tab:cont_type_example_3}.}
 \label{tab:contribution_type_prompt}
\end{table*}

\begin{table*}
\centering
\begin{tabular}{p{\dimexpr\linewidth-2\tabcolsep\relax}}
\hline
{\fontsize{9}{9}\selectfont \textbf{SYSTEM:}} \\
{\fontsize{9}{9}\selectfont Your task is to check whether the given paragraph, from a related work section draft for an academic paper, explicitly or implicitly mention the main paper's contribution or position among existing literature. In addition, some examples showing the implementation of the task will be provided. By utilizing the examples, first provide your reasoning, and then your answer as either \"yes\" or \"no\". Your output should be in JSON format.} \\ \\
{\fontsize{9}{9}\selectfont \textbf{USER:}} \\
{\fontsize{9}{9}\selectfont $<$START OF EXAMPLE 1$>$} \\
{\fontsize{9}{9}\selectfont \texttt{\{Example 1\}}} \\
{\fontsize{9}{9}\selectfont $<$END OF EXAMPLE 1$>$} \\ \\  
{\fontsize{9}{9}\selectfont $<$START OF EXAMPLE 2$>$} \\
{\fontsize{9}{9}\selectfont \texttt{\{Example 2\}}} \\
{\fontsize{9}{9}\selectfont $<$END OF EXAMPLE 2$>$} \\ \\
{\fontsize{9}{9}\selectfont DRAFT: \texttt{\{Paragraph from the generated related work section draft\}}} \\
\hline
\end{tabular}
\caption{Positioning ratio: System prompts and contrastive few-shot examples presented in Table \ref{tab:cont_direct_eval_examples} for positioning check for each paragraph.}
 \label{tab:cont_direct_eval_prompt}
\end{table*}

\begin{table*}
\centering
\begin{tabular}{p{\dimexpr\linewidth-2\tabcolsep\relax}}
\hline
{\fontsize{9}{9}\selectfont \textbf{SYSTEM:}} \\
{\fontsize{9}{9}\selectfont You will be given two paragraphs, context and final, from a related work section draft for an academic paper. Your task is to check whether the final paragraph states the main paper's contributions or its position among the literature while addressing the points from the context paragraph. In other words, you will check whether contributions in the final paragraph include the discussed points in the context paragraph. In addition, some examples showing the implementation of the task will be provided. By utilizing the examples, first provide your reasoning, and then your answer as either "yes" or "no". Your output should be in JSON format.} \\ \\
{\fontsize{9}{9}\selectfont \textbf{USER:}} \\
{\fontsize{9}{9}\selectfont $<$START OF EXAMPLE 1$>$} \\
{\fontsize{9}{9}\selectfont \texttt{\{Example 1\}}} \\
{\fontsize{9}{9}\selectfont $<$END OF EXAMPLE 1$>$} \\ \\  
{\fontsize{9}{9}\selectfont $<$START OF EXAMPLE 2$>$} \\
{\fontsize{9}{9}\selectfont \texttt{\{Example 2\}}} \\
{\fontsize{9}{9}\selectfont $<$END OF EXAMPLE 2$>$} \\ \\
{\fontsize{9}{9}\selectfont CONTEXT: \texttt{\{Context paragraph from the generated related work section draft\}}} \\ \\
{\fontsize{9}{9}\selectfont FINAL: \texttt{\{Final paragraph from the generated related work section draft\}}} \\
\hline
\end{tabular}
\caption{Positioning ratio: System prompts and contrastive few-shot examples  presented in Table \ref{tab:cont_pairwise_eval_examples} for positioning check for final paragraph comparisons with every other paragraph in the related work.}
 \label{tab:cont_pairwise_eval_prompt}
\end{table*}

\begin{table*}
\centering
\resizebox{\textwidth}{!}{
\begin{tabular}{l|cc|cc|cc|cc|cc}
\hline
\textbf{GPT-4o} & \multicolumn{2}{c|}{\textbf{1}} & \multicolumn{2}{c|}{\textbf{2}} & \multicolumn{2}{c|}{\textbf{3}} & \multicolumn{2}{c|}{\textbf{4}} & \multicolumn{2}{c}{\textbf{5}}\\ 
\textbf{Full} & \textbf{Mean} & \textbf{STD} & \textbf{Mean} & \textbf{STD} & \textbf{Mean} & \textbf{STD} & \textbf{Mean} & \textbf{STD} & \textbf{Mean} & \textbf{STD} \\ \hline
Hallucinated Papers & 0.97 & 0.05 & 0.97 & 0.05 & 1.0 & 0.0 & 1.0 & 0.0 & 1.0 & 0.0 \\ 
Missing Papers & 0.93 & 0.05 & 0.87 & 0.19 & 0.87 & 0.19 & 0.97 & 0.05 & 0.90 & 0.14 \\
Length & 0.0 & 0.0 & 0.10 & 0.09 & 0.0 & 0.0 & 0.10 & 0.09 & 0.07 & 0.09 \\
Citation Emphasis & 0.22 & 0.14 & 0.21 & 0.15 & 0.30 & 0.16 & 0.27 & 0.12 & 0.28 & 0.17 \\
Faithfulness & 0.17 & 0.19 & 0.10 & 0.14 & 0.10 & 0.14 & 0.13 & 0.14 & 0.13 & 0.19 \\
Positioning Existence & 1.0 & 0.0 & 1.0 & 0.0 & 1.0 & 0.0 & 1.0 & 0.0 & 1.0 & 0.0 \\
Positioning Type & 0.93 & 0.09 & 0.97 & 0.05 & 0.93 & 0.09 & 0.93 & 0.09 & 0.93 & 0.09 \\
Positioning Ratio & 0.93 & 0.09 & 0.97 & 0.05 & 0.93 & 0.09 & 0.93 & 0.11 & 0.93 & 0.09 \\ 
\hline
\end{tabular}
}
\caption{GPT-4o Full Pipeline results with mean and standard deviation (STD) across three runs at each iteration.}
\label{tab:gpt-full}
\end{table*}

\begin{table*}
\centering
\resizebox{\textwidth}{!}{
\begin{tabular}{l|cc|cc|cc|cc|cc}
\hline
\textbf{o3-mini} & \multicolumn{2}{c|}{\textbf{1}} & \multicolumn{2}{c|}{\textbf{2}} & \multicolumn{2}{c|}{\textbf{3}} & \multicolumn{2}{c|}{\textbf{4}} & \multicolumn{2}{c}{\textbf{5}}\\
\textbf{Full} & \textbf{Mean} & \textbf{STD} & \textbf{Mean} & \textbf{STD} & \textbf{Mean} & \textbf{STD} & \textbf{Mean} & \textbf{STD} & \textbf{Mean} & \textbf{STD} \\ \hline
Hallucinated Papers & 1.0 & 0.0 & 1.0 & 0.0 & 1.0 & 0.0 & 1.0 & 0.0 & 1.0 & 0.0 \\
Missing Papers & 0.97 & 0.05 & 0.83 & 0.09 & 1.0 & 0.0 & 0.77 & 0.14 & 0.97 & 0.05 \\
Length & 0.07 & 0.05 & 0.73 & 0.28 & 0.60 & 0.33 & 0.73 & 0.28 & 0.77 & 0.24 \\
Citation Emphasis & 0.25 & 0.12 & 0.19 & 0.15 & 0.29 & 0.16 & 0.25 & 0.14 & 0.25 & 0.13 \\
Faithfulness & 0.20 & 0.24 & 0.20 & 0.24 & 0.13 & 0.14 & 0.17 & 0.14 & 0.30 & 0.24 \\
Positioning Existence & 1.0 & 0.0 & 1.0 & 0.0 & 1.0 & 0.0 & 1.0 & 0.0 & 1.0 & 0.0 \\
Positioning Type & 1.0 & 0.0 & 1.0 & 0.0 & 0.97 & 0.05 & 1.0 & 0.0 & 1.0 & 0.0 \\
Positioning Ratio & 1.0 & 0.0 & 0.99 & 0.01 & 0.96 & 0.06 & 1.0 & 0.0 & 1.0 & 0.0 \\
\hline
\end{tabular}
}
\caption{o3-mini Full Pipeline results with mean and standard deviation (STD) across three runs at each iteration.}
\end{table*}

\begin{table*}
\centering
\resizebox{\textwidth}{!}{
\begin{tabular}{l|cc|cc|cc|cc|cc}
\hline
\textbf{Llama 3.3} & \multicolumn{2}{c|}{\textbf{1}} & \multicolumn{2}{c|}{\textbf{2}} & \multicolumn{2}{c|}{\textbf{3}} & \multicolumn{2}{c|}{\textbf{4}} & \multicolumn{2}{c}{\textbf{5}}\\
\textbf{Full} & \textbf{Mean} & \textbf{STD} & \textbf{Mean} & \textbf{STD} & \textbf{Mean} & \textbf{STD} & \textbf{Mean} & \textbf{STD} & \textbf{Mean} & \textbf{STD} \\ \hline
Hallucinated Papers & 0.80 & 0.28 & 0.93 & 0.09 & 0.97 & 0.05 & 1.0 & 0.0 & 1.0 & 0.0 \\
Missing Papers & 0.97 & 0.05 & 0.87 & 0.19 & 0.93 & 0.09 & 0.87 & 0.19 & 0.87 & 0.15 \\
Length & 0.0 & 0.0 & 0.23 & 0.24 & 0.13 & 0.09 & 0.17 & 0.24 & 0.27 & 0.24 \\
Citation Emphasis & 0.26 & 0.15 & 0.17 & 0.13 & 0.19 & 0.14 & 0.23 & 0.15 & 0.22 & 0.11 \\
Faithfulness & 0.17 & 0.19 & 0.17 & 0.19 & 0.0 & 0.0 & 0.0 & 0.0 & 0.07 & 0.09 \\
Positioning Existence & 0.97 & 0.05 & 1.0 & 0.0 & 1.0 & 0.0 & 1.0 & 0.0 & 1.0 & 0.0 \\
Positioning Type & 0.90 & 0.14 & 0.93 & 0.09 & 0.97 & 0.05 & 1.0 & 0.0 & 0.90 & 0.09 \\
Positioning Ratio & 0.59 & 0.30 & 0.74 & 0.20 & 0.79 & 0.17 & 0.77 & 0.18 & 0.73 & 0.19 \\ 
\hline
\end{tabular}
}
\caption{Llama 3.3 Full Pipeline results with mean and standard deviation (STD) across three runs at each iteration.}
\label{tab:llama-full}
\end{table*}

\begin{table*}
\centering
\resizebox{\textwidth}{!}{
\begin{tabular}{l|cc|cc|cc|cc|cc}
\hline
\textbf{Gemma 3} & \multicolumn{2}{c|}{\textbf{1}} & \multicolumn{2}{c|}{\textbf{2}} & \multicolumn{2}{c|}{\textbf{3}} & \multicolumn{2}{c|}{\textbf{4}} & \multicolumn{2}{c}{\textbf{5}}\\
\textbf{Full} & \textbf{Mean} & \textbf{STD} & \textbf{Mean} & \textbf{STD} & \textbf{Mean} & \textbf{STD} & \textbf{Mean} & \textbf{STD} & \textbf{Mean} & \textbf{STD} \\ \hline
Hallucinated Papers & 0.87 & 0.14 & 1.0 & 0.0 & 1.0 & 0.0 & 1.0 & 0.0 & 1.0 & 0.0 \\
Missing Papers & 0.70 & 0.24 & 0.67 & 0.33 & 0.63 & 0.33 & 0.53 & 0.33 & 0.63 & 0.24 \\
Length & 0.0 & 0.0 & 0.27 & 0.24 & 0.23 & 0.24 & 0.17 & 0.09 & 0.17 & 0.14 \\
Citation Emphasis & 0.21 & 0.09 & 0.25 & 0.11 & 0.24 & 0.11 & 0.27 & 0.11 & 0.30 & 0.11 \\
Faithfulness & 0.0 & 0.0 & 0.07 & 0.05 & 0.10 & 0.09 & 0.10 & 0.09 & 0.10 & 0.09 \\
Positioning Existence & 1.0 & 0.0 & 1.0 & 0.0 & 1.0 & 0.0 & 1.0 & 0.0 & 1.0 & 0.0 \\
Positioning Type & 1.0 & 0.0 & 1.0 & 0.0 & 0.97 & 0.05 & 1.0 & 0.0 & 1.0 & 0.0 \\
Positioning Ratio & 0.97 & 0.05 & 0.96 & 0.05 & 0.95 & 0.07 & 0.95 & 0.02 & 0.99 & 0.01 \\ 
\hline
\end{tabular}
}
\caption{Gemma 3 Full Pipeline results with mean and standard deviation (STD) across three runs at each iteration.}  
\end{table*}

\begin{table*}
\centering
\resizebox{\textwidth}{!}{
\begin{tabular}{l|cc|cc|cc|cc|cc}
\hline
\textbf{GPT-4o} & \multicolumn{2}{c|}{\textbf{1}} & \multicolumn{2}{c|}{\textbf{2}} & \multicolumn{2}{c|}{\textbf{3}} & \multicolumn{2}{c|}{\textbf{4}} & \multicolumn{2}{c}{\textbf{5}}\\
\textbf{New Paper} & \textbf{Mean} & \textbf{STD} & \textbf{Mean} & \textbf{STD} & \textbf{Mean} & \textbf{STD} & \textbf{Mean} & \textbf{STD} & \textbf{Mean} & \textbf{STD} \\ \hline
Hallucinated Papers & 1.0 & 0.0 & 1.0 & 0.0 & 1.0 & 0.0 & 1.0 & 0.0 & 1.0 & 0.0 \\
Missing Papers & 0.97 & 0.05 & 0.90 & 0.14 & 0.47 & 0.28 & 0.87 & 0.14 & 0.83 & 0.14 \\
Length & 0.0 & 0.0 & 0.10 & 0.09 & 0.03 & 0.05 & 0.03 & 0.05 & 0.03 & 0.05 \\
Citation Emphasis & 0.18 & 0.10 & 0.20 & 0.10 & 0.22 & 0.16 & 0.31 & 0.17 & 0.28 & 0.17 \\
Faithfulness & 0.13 & 0.19 & 0.10 & 0.14 & 0.13 & 0.14 & 0.10 & 0.14 & 0.07 & 0.09 \\
Positioning Existence & 1.0 & 0.0 & 1.0 & 0.0 & 1.0 & 0.0 & 1.0 & 0.0 & 1.0 & 0.0 \\
Positioning Type & 0.90 & 0.14 & 0.93 & 0.09 & 0.90 & 0.14 & 0.93 & 0.09 & 0.97 & 0.05 \\
Positioning Ratio & 0.89 & 0.14 & 0.92 & 0.10 & 0.87 & 0.16 & 0.93 & 0.09 & 0.94 & 0.07 \\ 
\hline
\end{tabular}
}
\caption{GPT-4o New Paper Pipeline results with mean and standard deviation (STD) across three runs at each iteration.}
\end{table*}

\begin{table*}
\centering
\resizebox{\textwidth}{!}{
\begin{tabular}{l|cc|cc|cc|cc|cc}
\hline
\textbf{o3-mini} & \multicolumn{2}{c|}{\textbf{1}} & \multicolumn{2}{c|}{\textbf{2}} & \multicolumn{2}{c|}{\textbf{3}} & \multicolumn{2}{c|}{\textbf{4}} & \multicolumn{2}{c}{\textbf{5}}\\
\textbf{New Paper} & \textbf{Mean} & \textbf{STD} & \textbf{Mean} & \textbf{STD} & \textbf{Mean} & \textbf{STD} & \textbf{Mean} & \textbf{STD} & \textbf{Mean} & \textbf{STD} \\ \hline
Hallucinated Papers & 1.0 & 0.0 & 1.0 & 0.0 & 1.0 & 0.0 & 1.0 & 0.0 & 1.0 & 0.0 \\
Missing Papers & 1.0 & 0.0 & 0.97 & 0.05 & 0.93 & 0.09 & 0.93 & 0.09 & 0.87 & 0.14 \\ 
Length & 0.13 & 0.14 & 0.87 & 0.09 & 0.53 & 0.28 & 0.73 & 0.24 & 0.77 & 0.19 \\ 
Citation Emphasis & 0.29 & 0.20 & 0.25 & 0.11 & 0.28 & 0.19 & 0.36 & 0.17 & 0.30 & 0.17 \\ 
Faithfulness & 0.20 & 0.24 & 0.27 & 0.28 & 0.27 & 0.24 & 0.27 & 0.24 & 0.20 & 0.14 \\ 
Positioning Existence & 1.0 & 0.0 & 1.0 & 0.0 & 1.0 & 0.0 & 1.0 & 0.0 & 1.0 & 0.0 \\ 
Positioning Type & 0.97 & 0.05 & 0.97 & 0.05 & 0.93 & 0.09 & 0.97 & 0.05 & 1.0 & 0.0 \\ 
Positioning Ratio & 0.95 & 0.08 & 0.97 & 0.05 & 0.93 & 0.09 & 0.97 & 0.05 & 1.0 & 0.0 \\ 
\hline
\end{tabular}
}
\caption{o3-mini New Paper Pipeline results with mean and standard deviation (STD) across three runs at each iteration.}
\end{table*}

\begin{table*}
\centering
\resizebox{\textwidth}{!}{
\begin{tabular}{l|cc|cc|cc|cc|cc}
\hline
\textbf{Llama 3.3} & \multicolumn{2}{c|}{\textbf{1}} & \multicolumn{2}{c|}{\textbf{2}} & \multicolumn{2}{c|}{\textbf{3}} & \multicolumn{2}{c|}{\textbf{4}} & \multicolumn{2}{c}{\textbf{5}}\\
\textbf{New Paper} & \textbf{Mean} & \textbf{STD} & \textbf{Mean} & \textbf{STD} & \textbf{Mean} & \textbf{STD} & \textbf{Mean} & \textbf{STD} & \textbf{Mean} & \textbf{STD} \\ \hline
Hallucinated Papers & 0.60 & 0.24 & 0.87 & 0.14 & 0.93 & 0.09 & 0.93 & 0.09 & 0.97 & 0.05 \\
Missing Papers & 0.93 & 0.09 & 0.90 & 0.0 & 0.77 & 0.33 & 0.97 & 0.05 & 0.93 & 0.05 \\ 
Length & 0.00 & 0.00 & 0.33 & 0.24 & 0.07 & 0.09 & 0.23 & 0.28 & 0.23 & 0.19 \\ 
Citation Emphasis & 0.18 & 0.15 & 0.22 & 0.12 & 0.28 & 0.12 & 0.33 & 0.18 & 0.41 & 0.22 \\ 
Faithfulness & 0.10 & 0.14 & 0.10 & 0.14 & 0.03 & 0.05 & 0.03 & 0.05 & 0.03 & 0.05 \\ 
Positioning Existence & 1.0 & 0.0 & 0.97 & 0.05 & 1.0 & 0.0 & 1.0 & 0.0 & 1.0 & 0.0 \\ 
Positioning Type & 0.87 & 0.14 & 0.90 & 0.09 & 0.90 & 0.09 & 0.90 & 0.09 & 0.93 & 0.05 \\ 
Positioning Ratio & 0.69 & 0.22 & 0.77 & 0.14 & 0.78 & 0.19 & 0.80 & 0.18 & 0.81 & 0.11 \\ 
\hline
\end{tabular}
}
\caption{Llama 3.3 New Paper Pipeline results with mean and standard deviation (STD) across three runs at each iteration.}
\end{table*}

\begin{table*}
\centering
\resizebox{\textwidth}{!}{
\begin{tabular}{l|cc|cc|cc|cc|cc}
\hline
\textbf{Gemma 3} & \multicolumn{2}{c|}{\textbf{1}} & \multicolumn{2}{c|}{\textbf{2}} & \multicolumn{2}{c|}{\textbf{3}} & \multicolumn{2}{c|}{\textbf{4}} & \multicolumn{2}{c}{\textbf{5}}\\
\textbf{New Paper} & \textbf{Mean} & \textbf{STD} & \textbf{Mean} & \textbf{STD} & \textbf{Mean} & \textbf{STD} & \textbf{Mean} & \textbf{STD} & \textbf{Mean} & \textbf{STD} \\ \hline
Hallucinated Papers & 0.57 & 0.18 & 0.97 & 0.05 & 1.0 & 0.0 & 1.0 & 0.0 & 1.0 & 0.0 \\ 
Missing Papers & 1.0 & 0.0 & 0.8 & 0.19 & 0.30 & 0.19 & 0.53 & 0.28 & 0.67 & 0.38 \\ 
Length & 0.0 & 0.0 & 0.40 & 0.14 & 0.13 & 0.14 & 0.07 & 0.09 & 0.30 & 0.24 \\ 
Citation Emphasis & 0.27 & 0.14 & 0.36 & 0.14 & 0.42 & 0.18 & 0.45 & 0.13 & 0.46 & 0.21 \\ 
Faithfulness & 0.0 & 0.0 & 0.03 & 0.05 & 0.07 & 0.05 & 0.0 & 0.0 & 0.0 & 0.0 \\ 
Positioning Existence & 1.0 & 0.0 & 1.0 & 0.0 & 1.0 & 0.0 & 1.0 & 0.0 & 1.0 & 0.0 \\ 
Positioning Type & 1.0 & 0.0 & 1.0 & 0.0 & 1.0 & 0.0 & 1.0 & 0.0 & 1.0 & 0.0 \\ 
Positioning Ratio & 0.99 & 0.02 & 0.96 & 0.05 & 0.99 & 0.01 & 0.99 & 0.01 & 0.95 & 0.06 \\ 
\hline
\end{tabular}
}
\caption{Gemma 3 New Paper Pipeline results with mean and standard deviation (STD) across three runs at each iteration.}
\end{table*}

\begin{table*}
\centering
\resizebox{\textwidth}{!}{
\begin{tabular}{l|cc|cc|cc|cc|cc}
\hline
\textbf{GPT-4o} & \multicolumn{2}{c|}{\textbf{1}} & \multicolumn{2}{c|}{\textbf{2}} & \multicolumn{2}{c|}{\textbf{3}} & \multicolumn{2}{c|}{\textbf{4}} & \multicolumn{2}{c}{\textbf{5}}\\ 
\textbf{Style Change} & \textbf{Mean} & \textbf{STD} & \textbf{Mean} & \textbf{STD} & \textbf{Mean} & \textbf{STD} & \textbf{Mean} & \textbf{STD} & \textbf{Mean} & \textbf{STD} \\ \hline
Hallucinated Papers & 1.0 & 0.0 & 1.0 & 0.0 & 1.0 & 0.0 & 1.0 & 0.0 & 1.0 & 0.0 \\ 
Missing Papers & 0.83 & 0.09 & 0.83 & 0.14 & 0.83 & 0.19 & 0.77 & 0.14 & 0.80 & 0.19 \\ 
Length & 0.07 & 0.05 & 0.0 & 0.0 & 0.03 & 0.05 & 0.07 & 0.05 & 0.03 & 0.05 \\ 
Citation Emphasis & 0.22 & 0.12 & 0.20 & 0.12 & 0.19 & 0.10 & 0.23 & 0.12 & 0.19 & 0.12 \\ 
Faithfulness & 0.17 & 0.24 & 0.17 & 0.19 & 0.13 & 0.19 & 0.17 & 0.24 & 0.07 & 0.09 \\ 
Positioning Existence & 1.0 & 0.0 & 1.0 & 0.0 & 1.0 & 0.0 & 1.0 & 0.0 & 1.0 & 0.0 \\ 
Positioning Type & 0.97 & 0.05 & 0.97 & 0.05 & 0.33 & 0.28 & 0.47 & 0.38 & 0.50 & 0.38 \\ 
Positioning Ratio & 0.96 & 0.06 & 0.95 & 0.07 & 0.33 & 0.28 & 0.47 & 0.38 & 0.49 & 0.37 \\ 
\hline
\end{tabular}
}
\caption{GPT-4o Style Change Pipeline results with mean and standard deviation (STD) across three runs at each iteration.}
\end{table*}

\begin{table*}
\centering
\resizebox{\textwidth}{!}{
\begin{tabular}{l|cc|cc|cc|cc|cc}
\hline
\textbf{o3-mini} & \multicolumn{2}{c|}{\textbf{1}} & \multicolumn{2}{c|}{\textbf{2}} & \multicolumn{2}{c|}{\textbf{3}} & \multicolumn{2}{c|}{\textbf{4}} & \multicolumn{2}{c}{\textbf{5}}\\ 
\textbf{Style Change} & \textbf{Mean} & \textbf{STD} & \textbf{Mean} & \textbf{STD} & \textbf{Mean} & \textbf{STD} & \textbf{Mean} & \textbf{STD} & \textbf{Mean} & \textbf{STD} \\ \hline
Hallucinated Papers & 1.0 & 0.0 & 1.0 & 0.0 & 1.0 & 0.0 & 1.0 & 0.0 & 1.0 & 0.0 \\ 
Missing Papers & 0.93 & 0.14 & 0.90 & 0.14 & 0.93 & 0.09 & 0.97 & 0.05 & 0.87 & 0.19 \\ 
Length & 0.03 & 0.05 & 0.70 & 0.33 & 0.53 & 0.28 & 0.67 & 0.28 & 0.63 & 0.28 \\ 
Citation Emphasis & 0.27 & 0.16 & 0.30 & 0.18 & 0.23 & 0.15 & 0.20 & 0.14 & 0.23 & 0.17 \\ 
Faithfulness & 0.20 & 0.14 & 0.33 & 0.28 & 0.10 & 0.09 & 0.13 & 0.05 & 0.27 & 0.19 \\ 
Positioning Existence & 1.0 & 0.0 & 1.0 & 0.0 & 1.0 & 0.0 & 1.0 & 0.0 & 1.0 & 0.0 \\ 
Positioning Type & 1.0 & 0.0 & 1.0 & 0.0 & 0.43 & 0.38 & 0.77 & 0.24 & 0.83 & 0.19 \\ 
Positioning Ratio & 0.99 & 0.01 & 0.99 & 0.01 & 0.42 & 0.37 & 0.77 & 0.24 & 0.83 & 0.19 \\ 
\hline
\end{tabular}
}
\caption{o3-mini Style Change Pipeline results with mean and standard deviation (STD) across three runs at each iteration.}
\end{table*}

\begin{table*}
\centering
\resizebox{\textwidth}{!}{
\begin{tabular}{l|cc|cc|cc|cc|cc}
\hline
\textbf{Llama 3.3} & \multicolumn{2}{c|}{\textbf{1}} & \multicolumn{2}{c|}{\textbf{2}} & \multicolumn{2}{c|}{\textbf{3}} & \multicolumn{2}{c|}{\textbf{4}} & \multicolumn{2}{c}{\textbf{5}}\\
\textbf{Style Change} & \textbf{Mean} & \textbf{STD} & \textbf{Mean} & \textbf{STD} & \textbf{Mean} & \textbf{STD} & \textbf{Mean} & \textbf{STD} & \textbf{Mean} & \textbf{STD} \\ \hline
Hallucinated Papers & 0.80 & 0.24 & 0.83 & 0.19 & 0.90 & 0.09 & 0.93 & 0.09 & 0.97 & 0.05 \\ 
Missing Papers & 0.93 & 0.09 & 0.77 & 0.28 & 0.83 & 0.14 & 0.87 & 0.14 & 0.83 & 0.24 \\ 
Length & 0.07 & 0.09 & 0.30 & 0.28 & 0.20 & 0.28 & 0.10 & 0.09 & 0.27 & 0.24 \\
Citation Emphasis & 0.23 & 0.09 & 0.18 & 0.10 & 0.28 & 0.17 & 0.25 & 0.11 & 0.20 & 0.12 \\
Faithfulness & 0.07 & 0.09 & 0.0 & 0.0 & 0.0 & 0.0 & 0.0 & 0.0 & 0.0 & 0.0 \\
Positioning Existence & 1.0 & 0.0 & 1.0 & 0.0 & 0.97 & 0.05 & 1.0 & 0.0 & 1.0 & 0.0 \\ 
Positioning Type & 0.93 & 0.09 & 0.77 & 0.19 & 0.30 & 0.33 & 0.60 & 0.33 & 0.53 & 0.47 \\
Positioning Ratio & 0.63 & 0.19 & 0.66 & 0.22 & 0.23 & 0.25 & 0.44 & 0.32 & 0.42 & 0.37 \\ 
\hline
\end{tabular}
}
\caption{Llama 3.3 Style Change Pipeline results with mean and standard deviation (STD) across three runs at each iteration.}
\end{table*}

\begin{table*}
\centering
\resizebox{\textwidth}{!}{
\begin{tabular}{l|cc|cc|cc|cc|cc}
\hline
\textbf{Gemma 3} & \multicolumn{2}{c|}{\textbf{1}} & \multicolumn{2}{c|}{\textbf{2}} & \multicolumn{2}{c|}{\textbf{3}} & \multicolumn{2}{c|}{\textbf{4}} & \multicolumn{2}{c}{\textbf{5}}\\
\textbf{Style Change} & \textbf{Mean} & \textbf{STD} & \textbf{Mean} & \textbf{STD} & \textbf{Mean} & \textbf{STD} & \textbf{Mean} & \textbf{STD} & \textbf{Mean} & \textbf{STD} \\ \hline
Hallucinated Papers & 0.87 & 0.05 & 0.97 & 0.05 & 1.0 & 0.0 & 1.0 & 0.0 & 1.0 & 0.0 \\
Missing Papers & 0.77 & 0.19 & 0.57 & 0.38 & 0.67 & 0.38 & 0.69 & 0.38 & 0.73 & 0.33 \\
Length & 0.0 & 0.0 & 0.20 & 0.19 & 0.20 & 0.24 & 0.23 & 0.33 & 0.27 & 0.14 \\ 
Citation Emphasis & 0.19 & 0.10 & 0.24 & 0.09 & 0.27 & 0.13 & 0.24 & 0.14 & 0.27 & 0.15 \\ 
Faithfulness & 0.07 & 0.09 & 0.03 & 0.05 & 0.03 & 0.05 & 0.03 & 0.05 & 0.07 & 0.09 \\ 
Positioning Existence & 1.0 & 0.0 & 1.0 & 0.0 & 1.0 & 0.0 & 1.0 & 0.0 & 1.0 & 0.0 \\ 
Positioning Type & 1.0 & 0.0 & 1.0 & 0.0 & 0.03 & 0.05 & 0.60 & 0.38 & 0.73 & 0.28 \\ 
Positioning Ratio & 0.94 & 0.06 & 0.96 & 0.06 & 0.03 & 0.05 & 0.56 & 0.36 & 0.70 & 0.29 \\ 
\hline
\end{tabular}
}
\caption{Gemma 3 Style Change Pipeline results with mean and standard deviation (STD) across three runs at each iteration.}
\label{tab:gemma-style}
\end{table*}

\end{document}